\documentclass{article}

\usepackage{microtype}
\usepackage{graphicx}
\usepackage{subfigure}
\usepackage{booktabs} 
\usepackage{svg}

\usepackage[linesnumbered,ruled,vlined,english,onelanguage]{algorithm2e}

\usepackage[accepted]{icml2024}

\usepackage{amsmath}
\usepackage{amssymb}
\usepackage{mathtools}
\usepackage{amsthm}

\usepackage{diagbox}
\usepackage{makecell}
\usepackage{xspace}
\usepackage{subcaption}
\usepackage{multirow}
\usepackage{wrapfig}
\usepackage{pifont}
\usepackage{xcolor,colortbl,pgf}
\usepackage[pagebackref=true,breaklinks=true,letterpaper=true,colorlinks,bookmarks=false,citecolor=green,linkcolor=red]{hyperref}\usepackage{url}            

\makeatletter 
  \newcommand\figcaption{\def\@captype{figure}\caption} 
  \newcommand\tabcaption{\def\@captype{table}\caption} 

\newcommand{\cmark}{\text{\ding{51}}}
\newcommand{\xmark}{\text{\ding{55}}}

\usepackage[capitalize,noabbrev]{cleveref}

\theoremstyle{plain}

\theoremstyle{definition}

\theoremstyle{remark}

\usepackage[textsize=tiny]{todonotes}

\icmltitlerunning{Efficient Training of Efficient GANs for Image-to-Image Translation}

\begin{document}

\twocolumn[
\icmltitle{E$^{2}$GAN: Efficient Training of Efficient GANs for Image-to-Image Translation}



\icmlsetsymbol{equal}{*}

\begin{icmlauthorlist}
\icmlauthor{Yifan Gong}{equal,comp,yyy}
\icmlauthor{Zheng Zhan}{equal,yyy}
\icmlauthor{Qing Jin}{comp}
\icmlauthor{Yanyu Li}{comp,yyy}
\icmlauthor{Yerlan Idelbayev}{yyy}
\icmlauthor{Xian Liu}{comp}
\icmlauthor{Andrey Zharkov}{comp}
\icmlauthor{Kfir Aberman}{comp}
\icmlauthor{Sergey Tulyakov}{comp}
\icmlauthor{Yanzhi Wang}{yyy}
\icmlauthor{Jian Ren}{comp}
\end{icmlauthorlist}
\centering
Project Page: \href{https://yifanfanfanfan.github.io/e2gan/}{https://yifanfanfanfan.github.io/e2gan/}
\icmlaffiliation{comp}{Snap Inc.}
\icmlaffiliation{yyy}{Northeastern University}

\icmlcorrespondingauthor{Jian Ren}{jren@snapchat.com}

\icmlkeywords{Machine Learning, ICML}

\vskip 0.3in
]



\printAffiliationsAndNotice{\icmlEqualContribution} 

\begin{abstract}
One highly promising direction for enabling flexible \emph{real-time} \emph{on-device} image editing is utilizing data distillation by leveraging large-scale text-to-image diffusion models to generate paired datasets used for training generative adversarial networks (GANs). 
This approach notably alleviates the stringent requirements typically imposed by high-end commercial GPUs for performing image editing with diffusion models. 
However, unlike text-to-image diffusion models, each distilled GAN is specialized for a specific image editing task, necessitating costly training efforts to obtain models for various concepts.
In this work, we introduce and address a novel research direction: \emph{can the process of distilling GANs from diffusion models be made significantly more efficient?} 
To achieve this goal, we propose a series of innovative techniques.
First, we construct a base GAN model with generalized features, adaptable to different concepts through fine-tuning, eliminating the need for training from scratch. Second, we identify crucial layers within the base GAN model and employ Low-Rank Adaptation (LoRA) with a simple yet effective rank search process, rather than fine-tuning the entire base model. Third, we investigate the minimal amount of data necessary for fine-tuning, further reducing the overall training time.
Extensive experiments show that we can efficiently empower GANs with the ability to perform real-time high-quality image editing on mobile devices with remarkably reduced training and storage costs for each concept.
\end{abstract}
    
\section{Introduction}
Recent development of diffusion-based image editing models has witnessed remarkable progress in synthesizing contents containing photo-realistic details full of imagination~\citep{saharia2022photorealistic,rombach2022high,ramesh2021zero,ramesh2022hierarchical}.
Albeit being creative and powerful, these generative models typically require a huge amount of computation even for inference and storage for saving weights.
For example, Stable Diffusion~\citep{rombach2022high} has more than one billion parameters and takes $30$ seconds to conduct an iterative denoising process to get one image on T4 GPU.
Such low-efficiency issue prohibits their real-time application on mobile devices~\citep{li2023snapfusion}.

Existing works try to tackle the problem through two main directions. One is accelerating the diffusion models by designing efficient model architecture or reducing the number of denoising steps~\citep{salimans2022progressive,meng2022distillation,li2022efficient,kim2023architectural}. However, these efforts still struggle to obtain models that can run in real-time on mobile devices~\citep{li2023snapfusion}. Another area focuses on data distillation, where diffusion models are leveraged to create datasets to train other mobile-friendly models, such as generative adversarial networks (GANs) for image-to-image translation~\citep{zhao2021large,parmar2023zero}. Nevertheless, although GAN is efficient for on-device deployment, each new concept still asks for the \emph{costly training} of a GAN model from \emph{scratch}.

In this work, we propose and aim to address a new research direction: \textit{can the GAN models be trained efficiently under the data distillation pipeline to perform real-time on-device image editing?} 

\begin{figure*} [t]
     \centering
     \includegraphics[width=1\textwidth]{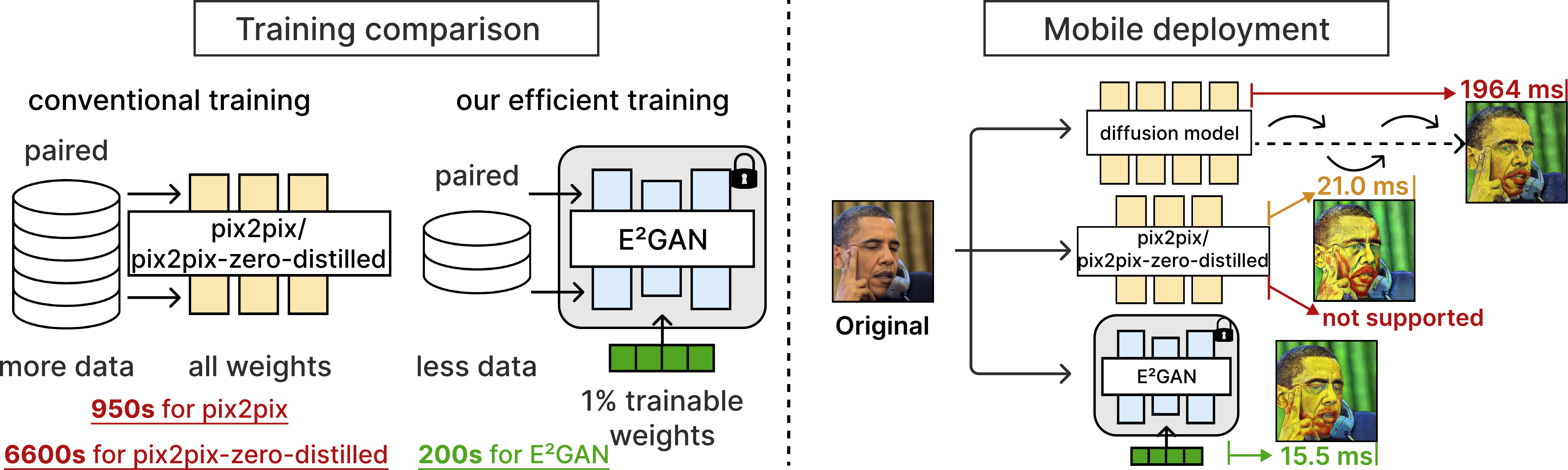}  
     \caption{\textbf{Overview of E$^2$GAN.} \emph{\textbf{Left}: Training Comparison.} Conventional GAN training, such as pix2pix~\citep{isola2017image} and pix2pix-zero-distilled that distills Co-Mod-GAN~\citep{zhao2021large} using data from a diffusion model~\citep{parmar2023zero}, requires all the weights trained from scratch, while our efficient training significantly reduces the training cost by only fine-tuning $1\%$ weights with only \emph{portion} of training data. \emph{\textbf{Right}: Mobile Inference Comparison.} Our efficient on-device model can achieve real-time ($30$FPS, iPhone 14) runtime and is faster than pix2pix and diffusion model, while the pix2pix-zero-distilled model (Co-Mod-GAN) is not supported on device.}
    \label{fig:motiv_images}  
\end{figure*}
To tackle the challenge, we introduce \textbf{E$^{2}$GAN}, powered with the following techniques for the \textbf{E}fficient training and \textbf{E}fficient inference of \textbf{GAN} models with the help of diffusion models:
\begin{itemize}
    \item First, we construct a base GAN model trained from various concepts and the corresponding edited images obtained from diffusion models.  It enables efficient transfer learning for different new concepts by fine-tuning, rather than training models from scratch, to reduce the training cost. Meanwhile, the base GAN model achieves fast inference with fewer parameters on mobile devices (as in Fig.~\ref{fig:motiv_images} \emph{Right}), and maintains high performance.
    \item Second, we identify that only partial layers are necessary to be fine-tuned for new concepts. LoRA is applied on these layers with a simple yet effective rank search process, eliminating the need to fine-tune the entire base model (as in Fig.~\ref{fig:motiv_images} \emph{Left}). It brings two advantages -- both the training cost and storage for each new concept are significantly reduced.
    \item Third, we investigate the amount of data for fine-tuning the base model for various concepts. Reducing the amount of training data helps reduce the training cost and time for adapting the base model to new concepts.
\end{itemize}

We show extensive experimental results to demonstrate that by using our approach, we can efficiently distill the image editing capability from a large-scale text-to-image diffusion model into GAN models via data distillation (examples in Fig.~\ref{fig:experiment_example}). The distilled GAN model showcases real-time image editing capabilities on mobile devices. We hope our work can shed light on how to democratize the diffusion models into efficient on-device computing.

\section{Related Works}
\textbf{Generative Models}. Generative models learn the joint data distribution to generate new samples, such as VAEs~\citep{kingma2013auto,rezende2014stochastic}, GANs~\citep{goodfellow2020generative,zhu2017unpaired,park2019SPADE}, auto-regressive models~\citep{van2016pixel,salimans2017pixelcnn++,van2016pixel,menick2018generating,yu2022scaling}, and diffusion models~\citep{sohl2015deep,ho2020denoising,nichol2021improved,song2020denoising,song2020score,dhariwal2021diffusion}.
Among these, diffusion models demonstrate a strong capability of generating images with high-fidelity~\citep{ramesh2022hierarchical,rombach2022high}, at the cost of bulky model size and numerous sampling steps during inference.
Several studies try to accelerate the image generation process of the diffusion models~\citep{salimans2022progressive,meng2022distillation,li2022efficient}. However, they still struggle to achieve real-time on-device generation.
On the contrary, GANs are more efficient in terms of model size and inference speed for image editing~\citep{li2020gan,jin2021teachers,wang2020gan}.
To this end, we leverage the approach of data distillation to transfer knowledge from diffusion models to lightweight GANs that are compatible with real-time inference on mobile devices.

\textbf{Efficient GANs.}
Existing works actively explore the reduction of the inference runtime for GANs by using various model compression techniques, such as efficient architecture design~\citep{li2020gan,jin2021teachers}, network pruning and quantization~\citep{wang2020gan,wang2019qgan}, and neural architecture search~\citep{wang2020gan,fu2020autogan}. For instance,  representative works like GAN Compression \cite{li2020gan} and GAN Slimming \cite{wang2020gan} mainly focus on the efficient model construction for the inference stage with reduced latency and model size, without considering the training cost. Specifically, GAN Compression \cite{li2020gan} decouples the model training and architecture search process for the obtaining of compressed weight values for inference, which leads to more computations during the training process.
On the other hand, the research about training cost savings for GANs is quite limited, as most works typically train all the parameters of a GAN model from scratch for the image-to-image translation task, involving large computing efforts. 
This work aims to fine-tune a very small portion, \ie $1\%$, of the pre-trained models with the partial training data to reduce the training cost. Thus, the training of GAN can be tiny in terms of both parameters and data.
There are many efforts on efficient training~\citep{huang2019gpipe, koster2017flexpoint}, in particular the sparse training~\citep{evci2020rigging, lee2018snip, yuan2022layer}.
However, these methods rely on the mask of trainable parameters, which in turn are determined during training with a huge bunch of data. In contrast, our method adopts pre-defined learnable components and only fine-tunes on a small fragment of data to make the transfer learning progress efficient and effective.

\begin{table}[]
    \centering
\caption{{\textbf{Comparison of model size, FLOPs, and latency} for different works~\citep{li2023snapfusion,isola2017image,parmar2023zero}.
Co-Mod-GAN ~\citep{zhao2021large} is trained following the pipeline in pix2pix-zero~\citep{parmar2023zero}. Reported latency is averaged over $100$ runs on iPhone 14 Pro. The training time of pix2pix and Co-Mod-GAN is measured on a single NVIDIA H100 GPU.} }\label{tab:size_flops_latency}
\scalebox{0.78}{
\begin{tabular}{l|cccc}
\Xhline{0.2ex}
 \textbf{Model} & \begin{tabular}[c]{@{}c@{}}     
      \textbf{Param} \\ 
      \textbf{num} 
      \end{tabular} & \textbf{FLOPs} & \textbf{Latency}  & \begin{tabular}[c]{@{}c@{}}     
      \textbf{Train} \\ 
      \textbf{time} 
      \end{tabular} \\ \hline
SnapFusion& 861M  & $>$1T & 1956 ms  & 7680 hours \\
Pix2pix with 9 RB & 11.4M & 56.9G & 21.0 ms & 16 min \\
Co-Mod-GAN  & 79.2M & 98.2G & not supported & 110 min \\
\Xhline{0.2ex}
\end{tabular}}
\end{table}

\section{Motivation}
 The huge model size, high computation cost, and numerous sampling steps pose significant challenges to the implementation of diffusion models on widely adopted mobile platforms with limited capacities. Even recent attempts at accelerating diffusion models, such as SnapFusion \cite{li2023snapfusion}, still require nearly 2 seconds to generate a single image on an iPhone 14 Pro, as shown in Tab. \ref{tab:size_flops_latency}. This efficiency issue strictly hinders their real-time application, \eg, image editing with $30$ frames per second (FPS), on widely adopted edge platforms such as mobile devices. 

{In contrast, various efficient and mobile-friendly GAN designs exist. For instance, the pix2pix model with 9 ResNet Blocks (RBs) takes only 21 ms to generate an edited image on an iPhone 14 Pro. Recognizing the inefficiency in directly accelerating diffusion models and the lightweight nature of certain GANs, researchers have explored data distillation as an alternative research direction. This approach involves transferring the knowledge of diffusion models to GANs. Latest work pix2pix-zero \cite{parmar2023zero} creates training data to train Co-Mod-GAN for model acceleration, yet it is not supported on mobile devices. Furthermore, the training time to obtain the Co-Mod-GAN for a new concept is still costly, which takes 110 min as shown in Tab. \ref{tab:size_flops_latency}.    }

{To overcome the above-mentioned limitations, the objective of this work is to achieve \textbf{efficient distillation} of diffusion models to \textbf{mobile-friendly real-time} GANs. Specifically, efficient distillation refers to minimizing the training efforts needed to obtain the GAN model for a new concept. Furthermore, when deployed on a mobile device after efficient distillation, the mobile-friendly real-time GANs should exhibit low latency ($<$33.3 ms) and demand minimal storage for a new concept.}
\section{Methods} 

In this section, we first give an overview of our knowledge transfer pipeline (Sec.~\ref{sec:knowledge_transfer}). 
{Then, we study efficient training strategies to get on-device models with \emph{reduced} training and storage costs, while maintaining high-quality image generation ability
(Sec.~\ref{sec:efficient_training}).}

\subsection{Overview of Knowledge Transfer Pipeline}\label{sec:knowledge_transfer}


\textbf{Pipeline for Dataset Creation.}
To enable the data distillation, we use the diffusion models to edit real images to obtain the edited images, forming pairs of data along with the used text prompts for the concept to create the training datasets, which can then be utilized to train the image-to-image GAN model. The real images come from FFHQ~\citep{karras2019style} and Flickr-Scenery~\citep{cheng2022inout}, covering diverse content and are challenging for content editing. 
For diffusion models, we choose the recent works for image editing, such as Stable Diffusion~\citep{rombach2022high}, Instruct-Pix2Pix (IP2P)~\citep{brooks2022instructpix2pix}, Null-text Inversion (NI)~\citep{mokady2022null}, ControlNet~\citep{zhang2023adding}, and InstructDiffusion~\citep{geng2023instructdiffusion}.
\begin{figure*} [htbp]
     \centering
\includegraphics[width=1.0\textwidth]{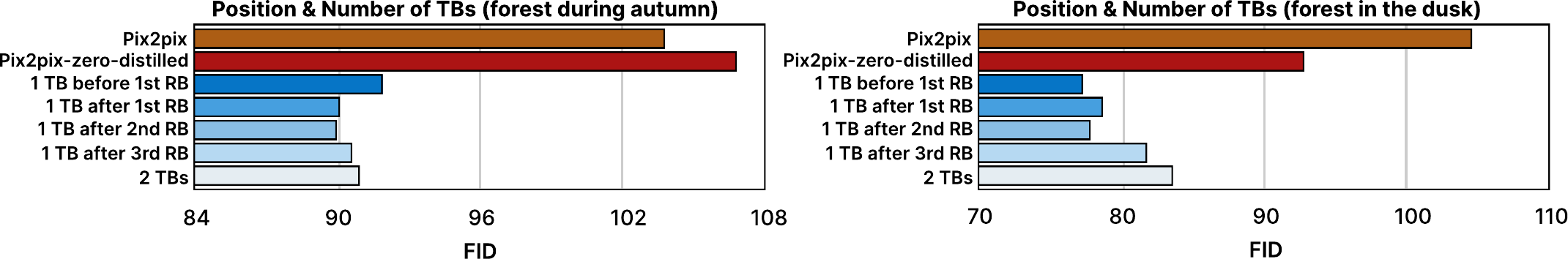}  
     \caption{ \textbf{FID comparison} of applying TBs in image generators trained on two datasets (\emph{Left:} \texttt{forest during autumn}, \emph{Right:} \texttt{forest in the dusk}).
     The vertical axis shows the position of inserting TBs. Pix2pix-zero-distilled uses pix2pix-zero for creating datasets to train Co-Mod-GAN~\citep{ramesh2021zero}.}
    \label{fig:model_architecture}
\end{figure*}
\begin{figure*} [t]
     \centering
     \includegraphics[width=1.0\textwidth]{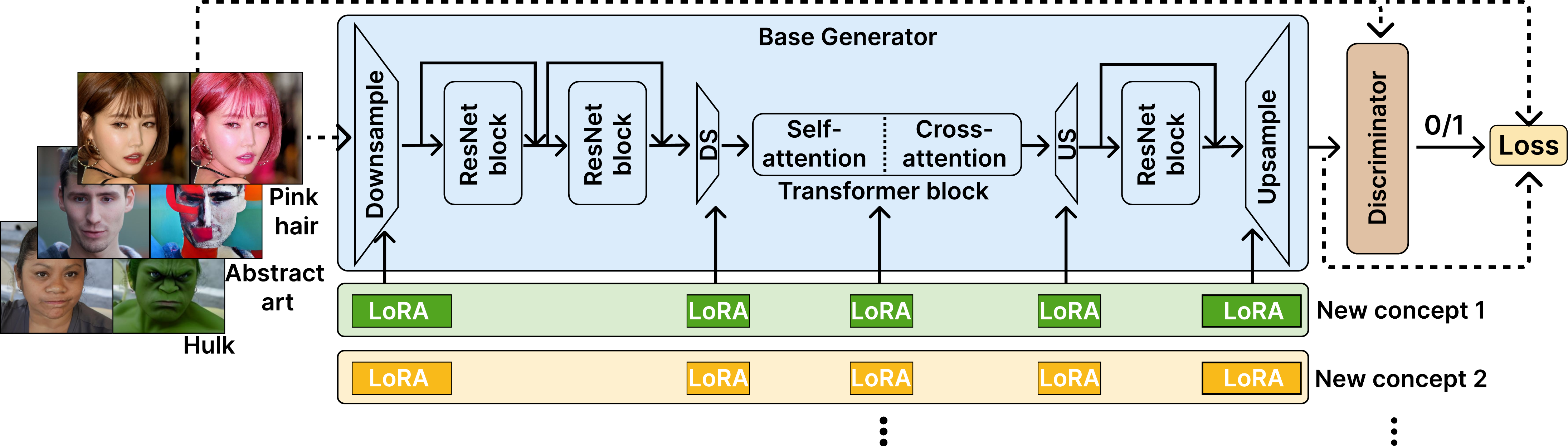}  
     \caption{\textbf{Overview of E$^2$GAN model architecture.} The generator is composed of down/up-sampling layers, 3 RBs, and 1 TB. The base generator is trained on multiple representative concepts. New concepts are achieved by fine-tuning LoRA parameters on crucial layers. }
    \label{fig:overview_arch}  
\end{figure*}

\textbf{Training Objectives.}
With paired images and the associated prompts for the concept, we train the efficient GANs for image translation by using the conventional adversarial loss.
Specifically, given the original image $\mathbf{x}$ and the editing prompt of the concept $\mathbf{c}$, the image generator $\mathcal{G}$ and discriminator $\mathcal{D}$ are jointly optimized as follows:
\begin{equation}\label{eq:loss} \scriptsize
\centering
\begin{aligned}
   &\min_{\theta_g} \max_{\theta_d} \lambda \underbrace{ \mathbb{E}_{\mathbf{x},\tilde{\mathbf{x}}^c,\mathbf{z}, \mathbf{c}} \left[ \| \tilde{\mathbf{x}}^c - \mathcal{G}(\mathbf{x}, \mathbf{z}, \mathbf{c};\theta_g) \|_1 \right]}_{\textrm{$\ell_1$ loss}} + \\ &\underbrace{\mathbb{E}_{\mathbf{x}, \tilde{\mathbf{x}}^c} 	\left[\log \mathcal{D} (\mathbf{x},\tilde{\mathbf{x}}^c; \theta_d) \right]  + \mathbb{E}_{\mathbf{x},{\mathbf{z}}, \mathbf{c}} 	\left[\log (1- \mathcal{D} (\mathbf{x}, \mathcal{G}(\mathbf{x},\mathbf{z}, \mathbf{c}; \theta_g); \theta_d)) \right]}_{\textrm{conditional GAN loss}},
   \end{aligned}
\end{equation}
where $\tilde{\mathbf{x}}^c$ denotes images generated by the diffusion model conditioned on the text prompt of the concept $\mathbf{c}$, $\mathcal{G}$ and $\mathcal{D}$ denote the generator and discriminator function parameterized by $\theta_g$ and $\theta_d$, respectively, $\mathbf{z}$ is a random noise introduced to increase the stochasticity of output, and $\lambda$ can be used to adjust the relative importance between two loss terms.

\subsection{Efficient Training of GAN Models}\label{sec:efficient_training}
Diffusion-based generative models can perform image editing on the fly while lightweight GAN-based networks typically require training to be adapted to the new concept. The training of GAN models for various concepts requires substantial computation costs. Additionally, there is a high storage demand for saving the trained weights.
To mitigate such training and storage costs, we introduce three main techniques to reduce the number of trainable parameters and the demanded data for model fine-tuning: \emph{First}, we establish a \emph{base GAN model} equipped with generalized features and representations, ready to be leveraged for new concepts (Sec.~\ref{sec:base_model_con}). \emph{Second}, starting from the base model, we identify key parameters to optimize during fine-tuning for a new concept, bolstered by the application of LoRA~\citep{hu2021lora} to further reduce the number of parameters (Sec.~\ref{sec:crucial_weights}).
\emph{Third}, we explore the possibility of tiny fine-tuning where the training data are first clustered and only those near the cluster centers are used (Sec.~\ref{sec:sc}).

\subsubsection{Base GAN Model Construction}\label{sec:base_model_con}
To obtain model weights for a new target concept with as few training efforts as possible, we explore transfer learning from a pre-trained base GAN model, instead of training from scratch. The base model should possess the capability of more general features and representations, which can be learned from multiple image translation tasks, allowing the new concept to leverage existing knowledge. Thus, we opt to train the base model on a mixed dataset comprising diverse concepts. 

{The construction of the image-to-image model $\mathcal{G}$ serves as the first step in obtaining such a base model.  This model should fulfill three key criteria: (1) the ability to learn multiple concepts; (2) achievement of real-time inference on mobile devices; and (3) strong image generation capabilities.  We start from the classic ResNet generator with $9$ RBs that is widely adopted~\citep{isola2017image,zhu2017unpaired,park2020contrastive}. To incorporate the text information of the concept and facilitate a more holistic understanding of global shapes and structure, we introduce Transformer Blocks (TBs) with self-attention and cross-attention modules into the architecture. For expedited inference purposes, we reduce the number of RBs from $9$ to $3$. The subsequent steps involve determining the number and position of TBs.}

\begin{table}[]
    \centering
 \caption{The model size, FLOPs, and latency of E$^2$GAN. The reported latency is an average of 100 runs measured on the GPU of an iPhone 14 Pro. } \label{tab:e2gan_size}
\scalebox{1}{
\begin{tabular}{l|ccc}
\Xhline{0.2ex}
\textbf{Model} & \textbf{Param num} & \textbf{FLOPs} & \textbf{Latency}  \\ \hline
 \rowcolor[gray]{.9} \textbf{3RB+1TB} & \textbf{7.1M} & \textbf{23.6G} & \textbf{15.5 ms}  \\
3RB+2TB & 10.1M & 26.6G & 21.0 ms \\ \Xhline{0.2ex}
\end{tabular}}
\end{table}

{\textbf{Number of TBs.} We train models with different architecture designs, \eg different numbers of TBs, and evaluate both the efficiency (in terms of model size, FLOPs, and latency) and image generation capability (in terms of the FID~\citep{heusel2017gans} between the images generated by GANs and diffusion models). The results are presented in Tab. \ref{tab:e2gan_size} and Fig.~\ref{fig:model_architecture}, respectively. Interestingly, we find that one TB is enough to generate high-quality images. Introducing more TBs does not further improve the performance yet brings in more computation cost. Notice that to reduce the inference cost of the introduced TB, we apply a downsampling operation to halve the feature map size before sending it into the TB, and use an upsampling layer to recover the feature map size for the following operations.  }


\begin{figure}[]
\centering{
\includegraphics[width=1.0 \linewidth]{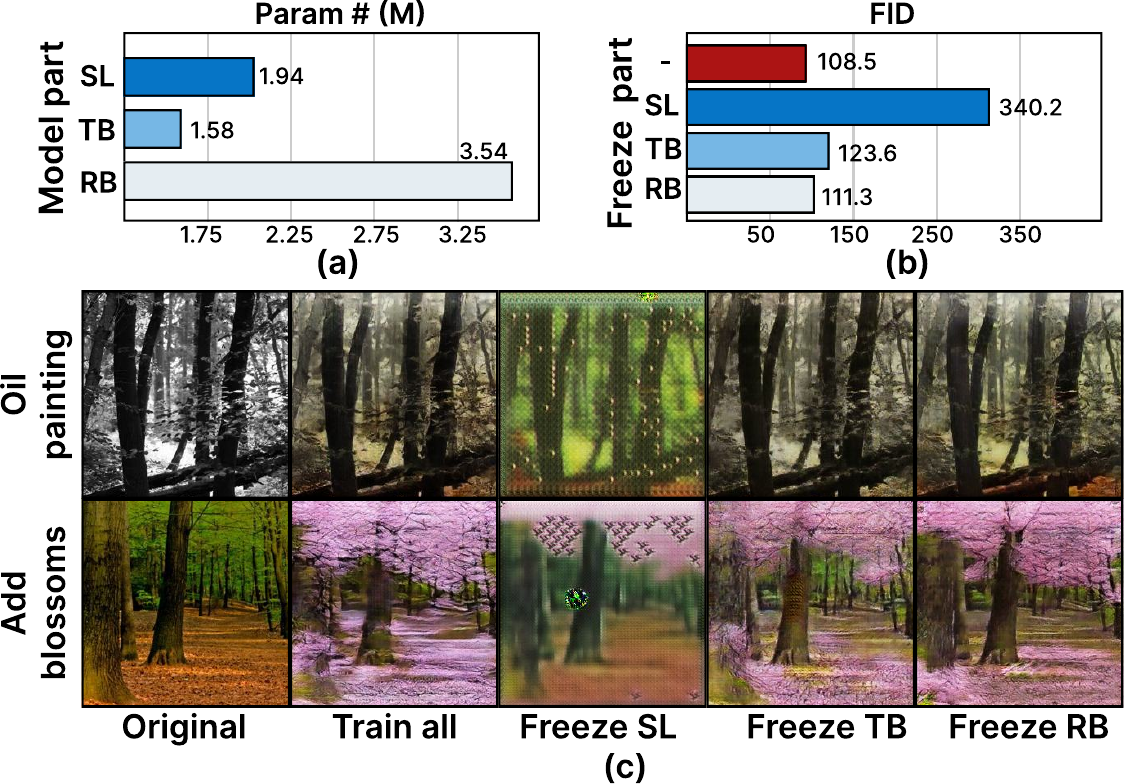}
}
\caption{{\textbf{Crucial weights analysis} via freezing partial weights in the base model. \textbf{(a)} Number of parameters for each part of the base model; \textbf{(b)} Averaged FID across $10$ different concepts on the Flicker-Scenery dataset when freezing partial weights of base model. `-' indicates fine-tuning all the weights; \textbf{(c)} The generated images when freezing each part of the base model.}
}
\label{fig:crucial_weight} 
\end{figure}
\textbf{Position of TBs.} Additionally, we find that the position of the TB is important for the final performance of the image generation. First, the TB should be placed between the last downsample layers and the first upsample layers to avoid high computations on mobile devices, due to the high resolution of features. 
Second, we apply attention to different positions of the network bottleneck. Particularly, the TB can be inserted between one of the following: (1) before the first RB; (2) after the first RB; (3) after the second RB; and (4) after the third RB. 
As evident in Fig.~\ref{fig:model_architecture}, all these options lead to a generator with better performance than the conventional CONV-only networks used in pix2pix~\citep{isola2017image} and pix2pix-zero-distilled~\citep{parmar2023zero}. For our model, we place the TB after the second RB.

Thus,  our architecture is finalized with an overall architecture in Fig.~\ref{fig:overview_arch}. It achieves faster inference speeds, reduces the number of parameters, and lowers computational costs compared to existing image-to-image models, as shown in Tab. \ref{tab:e2gan_size} and Fig.~\ref{fig:model_architecture}. With the architecture determined, the base model is trained on a subset of concepts denoted as $\mathcal{C}=\{\mathbf{c}_1,\cdots, \mathbf{c}_K\}$, where each concept $\mathbf{c}_k$ is selected among different concepts by K-means clustering~\citep{lloyd1982least} based on the average of the CLIP image embedding~\citep{radford2021learning} of uniformly sampled images.

\subsubsection{Crucial Weights for Fine-Tuning} \label{sec:crucial_weights}
To save the training and storage costs, we reduce the number of trainable parameters during fine-tuning.
Specifically, we pre-define trainable layers that occupy a small portion of weights from the base model. Then, we apply LoRA on top of the trainable layers.
In this way,  we only optimize $1.29\%$ of the weights from the base model during fine-tuning, greatly reducing the training and storage costs for a new concept.


Inspired by the recent work of customized diffusion~\citep{kumari2022multi}, which demonstrates that a pre-trained diffusion model can be fine-tuned to a personalized version by updating only a subset of its weights, we explore the feasibility of identifying the minimal set of tunable weights for GANs. Our objective is to determine a set of weights that is sufficient for fine-tuning the base model to adapt to a new concept.
To this end, we analyze the components of the GAN model, which mainly consist of three parts: (1) sampling layers (SL) with downsampling and upsampling; (2) transformer block (TB); and (3) intermediate RB.

\textbf{Identifying Crucial Layers.}
We systematically and empirically study the impact of each part in the image-to-image task by freezing each part in the model individually, with results provided in Fig.~\ref{fig:crucial_weight}. Combining  Fig.~\ref{fig:crucial_weight}(b) \& (c), we see that SL plays a more crucial role in maintaining the quality of generated images, identified by the high FID score value and low image quality. 
SL might be more crucial for constructing the desired output texture, yet intermediate RB might contain lower-level information that are common among styles. Meanwhile, compared to RB, TB has a fewer amount of parameters ($1.58$M \emph{v.s.} $3.54$M in Fig.~\ref{fig:crucial_weight}(a)), while it is more important in keeping performance ($123.6$ \emph{v.s.} $111.3$ in Fig.~\ref{fig:crucial_weight}(b)). Considering the situation with a limited training budget, RB has a lower priority to be optimized.  



\textbf{LoRA on Crucial Layers.}
From the perspective of maintaining image-generating quality, it is better to include TB in training as self-attention modifies the image with a better holistic understanding and the cross-attention module takes the information from the given target concept. However, combining SL and TB leads to $3.42$M parameters to be updated, taking up $47.90\%$ of the entire model weights. To fine-tune the crucial layers with much fewer trainable parameters, we investigate the best way of incorporating Low-Rank Adaptation (LoRA)~\citep{hu2021lora} into GAN training, which introduces two trainable low-rank weight matrices besides the original weight for each layer identified as crucial. By doing so, not only the training efforts, but also the storage costs for a new concept are significantly reduced.

\textbf{Rank for LoRA.} \label{sec:rank_for_lora}
With the leverage of LoRA, when fine-tuning to a new concept, the weights of the base model are \emph{frozen}, while only the two low-rank matrices with much fewer parameters for each crucial layer are updated to save computation and storage costs. For instance, for a CONV layer $i$ with weights $\small \theta_i \in \mathbb{R}^{h\times w \times
k_h\times k_w}$,  we apply two low-rank matrices with rank $r_i$, \ie $\small \theta_{i}^A \in \mathbb{R}^{h\times r_i \times k_h \times k_w}$ and  $\small \theta_{i}^B \in \mathbb{R}^{r_i \times w \times 1\times1}$, to approximate the gradient update $\nabla \theta_i$. Given multiple crucial layers, determining the appropriate rank for \emph{each of them} is important. Prior works mostly rely on manual setting \citep{hu2021lora} for deciding the rank value, due to a huge search space for the rank. 
However, in our task, the rank should be pre-fixed for different concepts to avoid the rank search process when a new concept comes. To tackle this challenge, we randomly sample $K$ concepts and conduct a simple yet effective rank search process. For each concept, we start by assigning $r_i$ as 1 for each crucial layer $i$, and upscale the rank for every $e$ epochs by doubling the rank value, until $r_i$ reaches the upper threshold $\tau_i$ for the layer $i$. The threshold $\tau_i$ is determined by the size of the weight. We evaluate the FID performance at the end of each $e$ training epochs. If the performance saturates, the rank value from the best FID performance setting is returned as the rank for the concept. Typically, a larger rank can provide more model capability. Thus, the largest returned rank among the $K$ selected concepts is viewed as $r^*$ for the future use of a new concept. The overall algorithm is described in Algorithm \ref{alg:lora_search} in Sec. \ref{appendix:algorithim} in the Appendix.

\begin{figure*} [t]
     \centering
     \includegraphics[width=0.91\textwidth]{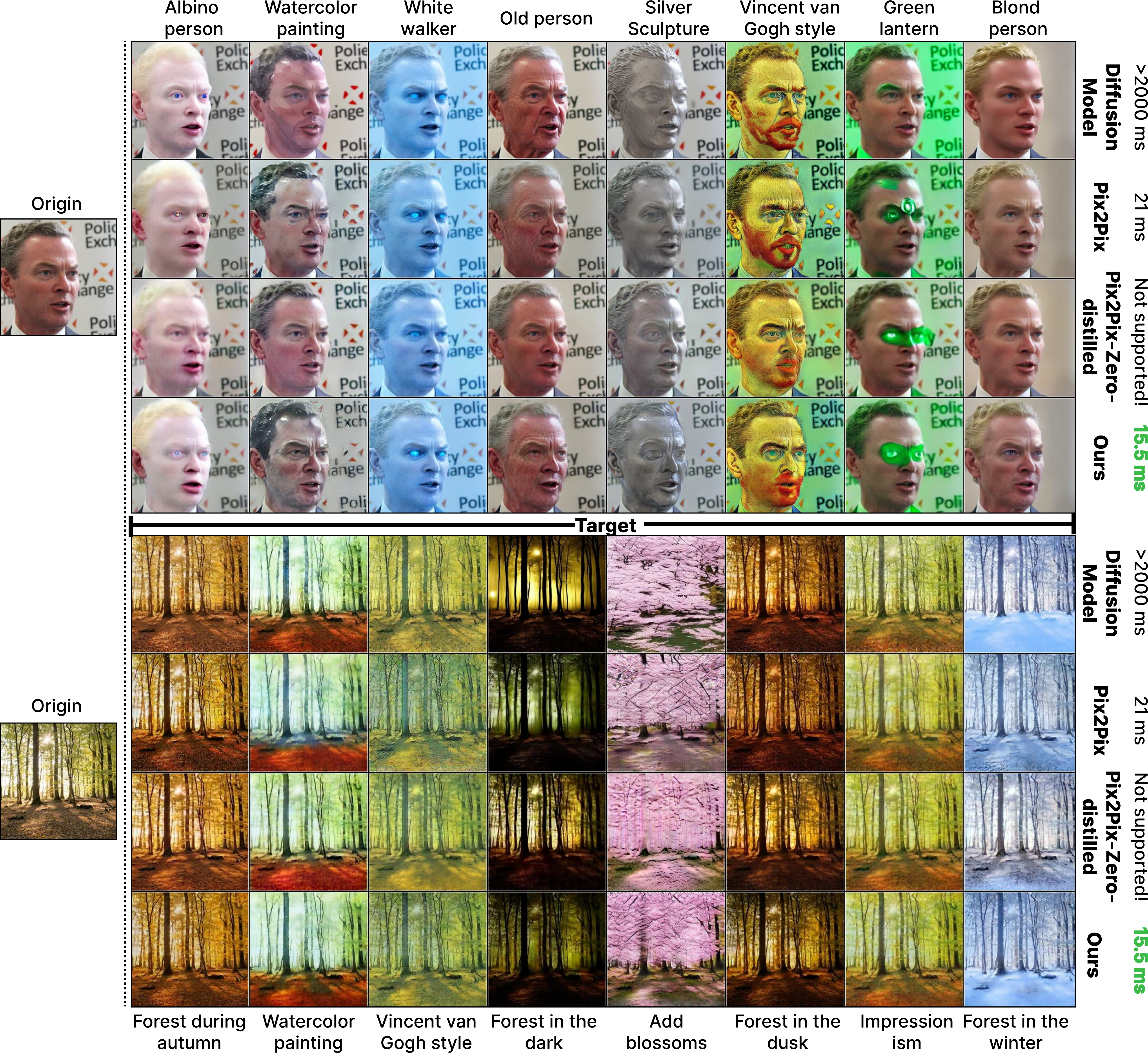}  
     \caption{{\textbf{Qualitative comparisons} on various tasks. The \emph{leftmost} column shows two original images and the remaining columns present the corresponding synthesized images in the target concept domain, where target prompts are shown at the bottom row. We provide images generated by various models.}}
    \label{fig:experiment_example}
\end{figure*}
\subsubsection{Training Data Reduction} \label{sec:sc}

Reducing the amount of training data can directly result in a reduction in the training time. Thus, we aim to investigate data efficiency as a means of decreasing the training cost in addition to the crucial weight update for E$^{2}$GAN.
We find not all data are indispensable for reliable training, but only a small subset is necessary. We obtain this small subset in an unsupervised manner with a selection of the data crowding around the clustering center on the whole dataset. 

{To identify the small subset of essential data, we conduct unsupervised learning to analyze the structure of the training data.
We first extract an embedding 
for each image $\mathbf{x}$ with an extractor $\mathcal{E}$. Then, we apply clustering on the embeddings by the K-Means algorithm~\citep{lloyd1982least} to obtain $K<N$ clusters ($N$ is the total number of training images), each with center $\mu_k$. The embeddings within the same cluster have a closer distance from each other, indicating a higher \emph{similarity} of the data points. To reduce the data amount while maintaining data diversity for the good generalization ability of the model, one data point, which is the closest to the center $\mu_k$, is selected for each of the $K$ clusters.

With our data selection method using $K$ clusters, we further reduce the number of training iterations by $N/K$ times.
In contrast to prior methods involving additional computations in the training process to shrink the dataset~\citep{yuan2021mest,wang2022sparcl}, our Similarity Clustering (SC) data reduction is tailored for expediting the training of image editing tasks.
It reduces the training data volume directly before the training process without incurring any additional costs during the training.

\section{Experiments}

In this section, we provide the detailed experimental settings and results of our proposed method. More details as well as some ablation studies can be found in the Appendix. 

\subsection{Experiments Setup}
\textbf{Paired Data Preparation.} We verify our method on $1,000$ images from FFHQ dataset~\citep{karras2019style} and Flickr-Scenery dataset~\citep{cheng2022inout} with image resolution as $256\times256$. The images in the target domain are generated with several different text-to-image diffusion models, including Stable Diffusion~\citep{rombach2022high}, IP2P~\citep{brooks2022instructpix2pix},  NI~\citep{mokady2022null}, ControlNet~\citep{zhang2023adding}, and InstructDiffusion~\citep{geng2023instructdiffusion}. The generated images with the best perceptual quality among diffusion models are selected to form with the real images into paired datasets. 
To perform training and evaluation of GAN models, we divide the image pairs from each target concept into training/validation/test subsets with the ratio as $80\%$/$10\%$/$10\%$. All the concepts to evaluate for the fine-tuning performance are reserved from the other concepts. 

\textbf{Baselines.} We compare E$^{2}$GAN with image-to-image translation methods like pix2pix~\citep{isola2017image} (image generator with $9$ ResNet blocks) and pix2pix-zero-distilled that distills Co-Mod-GAN~\citep{zhao2021large} using data generated by pix2pix-zero~\citep{parmar2023zero}.

\textbf{Training Setting.} We follow the standard approach that alternatively updates the generator and discriminator~\citep{goodfellow2020generative}.
The training is conducted from an initial learning rate of $2e-4$ with mini-batch SGD using Adam solver~\citep{kingma2014adam}.
The total training epochs is set to $100$ for E$^{2}$GAN, and $200$ for pix2pix~\citep{isola2017image} and pix2pix-zero-distilled~\citep{parmar2023zero} for them to converge well.
For SC (Sec.~\ref{sec:sc}), we choose the cluster number as $400$ and use the feature extractor $\mathcal{E}$ as FaceNet~\citep{schroff2015facenet} on FFHQ dataset and CLIP image encoder~\citep{radford2021learning} on Flicker Scenery dataset.
To train the base model, we use $20$ prepared tasks/datasets from the FFHQ dataset and $7$ from the Flickr Scenery dataset. The training and training time measurements are conducted on one NVIDIA H100 GPU with 80 GB memory.   

\textbf{Evaluation Metric.} We compare the images generated by E$^2$GAN and baseline methods by calculating Clean FID proposed by~\cite{parmar2022aliased} on the test sets. 


\subsection{Experimental Results}\label{sec:overall_performance}

\textbf{Qualitative Results.}
The synthesized images in the target domain obtained by E$^2$GAN and other methods are shown in Fig.~\ref{fig:experiment_example}. The original images are listed at the leftmost column, and the synthesized images for the target concept obtained by diffusion models, pix2pix, pix2pix-zero-distilled, and E$^2$GAN are shown from top to bottom.
The tasks span a wide range, such as changing the age, artistic styles, and editing the seasons.
According to the results, E$^2$GAN is able to modify the original images to the target concept domain by updating only the LoRA parameters. For instance, for the \texttt{green lantern} concept on the FFHQ dataset, the diffusion model fails to modify the image, pix2pix and pix2pix-zero-distilled add colors to wrong areas, while E$^2$GAN generates the image that fits the concept well. As for the \texttt{add blossoms} concept on the Flicker Scenery dataset, E$^2$GAN preserves the structure of the original image better than other models while editing the image as desired.
\begin{figure} []
     \centering
    \includegraphics[width=1.0\linewidth]{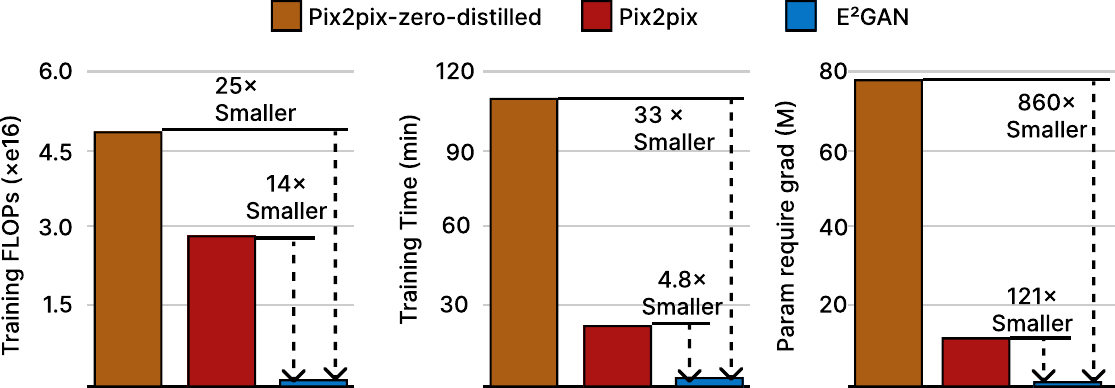}
 \caption{ \textbf{Training cost comparison} of baselines and E$^2$GAN. \textbf{\emph{Left}:} Training FLOPs. \textbf{\emph{Middle}:} Training time. \textbf{\emph{Right}:} Number of parameters that require gradient update (equals the weights that need to be saved for a concept).} \label{fig:train_cost}
\end{figure}
\begin{table}[]
\centering
\caption{ \textbf{FID comparison.} The FID is calculated between the images generated by GAN-based approaches and diffusion models.
Reported FID is averaged across different concepts ($30$ for FFHQ and $10$ for Flicker Scenery).} \label{tab:comparison}
\scalebox{1.0}{
\begin{tabular}{l|ccc}
\Xhline{0.2ex}

\diagbox{\textbf{Method}}{\textbf{Dataset}} &   FFHQ   & Landscape \\ \hline 
                      
Pix2pix                & 86.03  & 114.2     \\
Pix2pix-zero-distilled & 87.76  & 132.6     \\
\rowcolor[gray]{.9} \textbf{E$^{2}$GAN}             & \textbf{80.28} & \textbf{109.37} \\
\Xhline{0.2ex}
\end{tabular}
}
\end{table}

\textbf{Quantitative Comparisons}. The quantitative comparisons between E$^2$GAN and other baseline methods on the two datasets are provided in Tab. \ref{tab:comparison}. Note that for each concept, pix2pix and pix2pix-zero-distilled are trained on the whole training dataset of $800$ samples. E$^2$GAN begins with a base model and is fine-tuned with only $400$ samples on LoRA weights to obtain models for different target concepts. 
The results demonstrate that E$^2$GAN can reach an even better FID performance than the conventional GAN training techniques like pix2pix and pix2pix-zero-distilled, indicating high-fidelity of generated images.

\textbf{Training Cost Analysis}. We show the training cost comparisons between E$^2$GAN and other approaches in Fig.~\ref{fig:train_cost} in terms of training FLOPs, training time, and number of parameters that require gradient update.
Compared with pix2pix and pix2pix-zero-distilled,
E$^2$GAN greatly saves the training FLOPs of $14\times$ and $25\times$, respectively, and accelerates the training time by $4.8\times$ and $33\times$, respectively.
Moreover, E$^2$GAN only requires updating $0.092$M parameters for a new concept, greatly saving the storage requirement when training models for various tasks/concepts, \ie $869\times$ less than pix2pix-zero-distilled. 

Notably, E$^2$GAN requires \emph{much fewer} trainable parameters, training data, and training time than other GAN-based approaches to reach even \emph{better} generation quality, \ie E$^2$GAN has lower FID than pix2pix on FFHQ ($80.28$ \emph{v.s.} $86.03$). Furthermore, E$^2$GAN enjoys a faster inference speed on mobile devices (Tab.~\ref{tab:size_flops_latency}).
The good performance of E$^2$GAN originates from our effective framework design, including the efficient model architecture and efficient training strategy that reduces the training parameters and training data (Sec.~\ref{sec:efficient_training}). The results showcase the possibility of democratizing the powerful diffusion models into efficient on-device computing.

\subsection{Ablation Analysis}\label{sec:ablation_analysis}
We provide ablation analysis to understand the impact of each component in our efficient GAN training pipeline. We first study the effectiveness of the base model determination.
After that, we provide an analysis of the LoRA rank search.
Finally, we discuss the effect of our data selection. 

\begin{table}[]
\centering
\caption{ \textbf{Analysis (FID) of various base models} on FFHQ. } \label{tab:base_model_influence}
\scalebox{0.8}{
\begin{tabular}{c|cccc}
\Xhline{0.2ex}
\diagbox{\textbf{Concept}}{\textbf{Base model}}       & \cellcolor[gray]{.9}\textbf{Ours}  & \begin{tabular}[c]{@{}c@{}}     
      20 \\ 
      random 
      \end{tabular} & \begin{tabular}[c]{@{}c@{}}     
      200 art \\ 
      concepts 
      \end{tabular} &   \begin{tabular}[c]{@{}c@{}}     
       Single \\ 
      concept
      \end{tabular}\\ \hline
\texttt{White walker}  & \cellcolor[gray]{.9} \textbf{40.18} & 53.92     & 40.32 & 51.99         \\ \hline
\texttt{Blond person}  & \cellcolor[gray]{.9} \textbf{48.01} & 52.77     & 61.50        & 55.58  \\ \hline
\texttt{Sunglasses}    & \cellcolor[gray]{.9} \textbf{38.49} & 40.54     & 41.37  & 44.12         \\ \hline
\texttt{Vangogh style} & \cellcolor[gray]{.9} 71.82 & 78.58     & \textbf{68.21} & 78.06         \\\Xhline{0.2ex}
\end{tabular}
}
\end{table}
\textbf{Analysis of Base Model Determination.}
We study the impacts of our base model determination method discussed in Sec. \ref{sec:base_model_con} by comparing our method with the following three settings:
(1) train the base model on $20$ \emph{random} concepts; (2) train the base model on $200$ artist concepts; (3) train the base model on a \emph{single} concept \texttt{old person} from the FFHQ dataset.  The results are demonstrated in Tab.~\ref{tab:base_model_influence}. Note our method is obtained by training on $20$ selected representative concepts. The results indicate our base model construction outperforms or matches the alternatives across the evaluated concepts. This underscores the efficacy of our base model in enhancing performance. In contrast, the single concept base model generally performs worse. Furthermore, simply increasing the amount of concepts does not necessarily lead to better performance as indicated by training the base model with $200$ art concepts.

\begin{table}[]
\centering
\caption{\textbf{Analysis of searching  LoRA rank} on the Flickr Scenery dataset. The reported FID values are averaged over $10$ different target concepts.} \label{tab:wo_lora}
\scalebox{0.9}{
\begin{tabular}{c|cc}
\Xhline{0.2ex}
\textbf{Scheme}          & \textbf{FID}    & \textbf{\# of Param} \\ \hline
\rowcolor[gray]{.9} \textbf{Our searched}       & \textbf{109.37} & 0.092M    \\
Upscale 1$\times$ & 130.98 & 0.056M    \\
Upscale 4$\times$ & 111.42 & 0.164M    \\
Random          & 129.87 & 0.100M    \\ \Xhline{0.2ex}
\end{tabular}
}
\end{table}
\textbf{Analysis of LoRA on Crucial Layers.}
Tab.~\ref{tab:wo_lora} presents an evaluation of the effectiveness of our LoRA rank search on the Flicker Scenery dataset. The table reports the FID averaged across $10$ different target concepts, as well as the number of LoRA parameters for various schemes. We compare our method with the other three settings: (1) upscale the rank $1\times$ for each crucial layer by doubling the rank from the initialization until the rank reaches the threshold; (2) upscale the rank $4\times$ for each crucial layer from the initialization; and (3) random assign ranks for the crucial layers.  The results indicate that our searched scheme achieves the lowest FID value of $109.37$ while maintaining a relatively low number of parameters as $0.092$M. Though settings (2) and (3) use more parameters for fine-tuning, the FID performance is worse than our searched scheme. This demonstrates the 
importance of the appropriate rank setting and the effectiveness of our LoRA rank search approach. 

\begin{figure} []
     \centering
\includegraphics[width=0.92\linewidth]{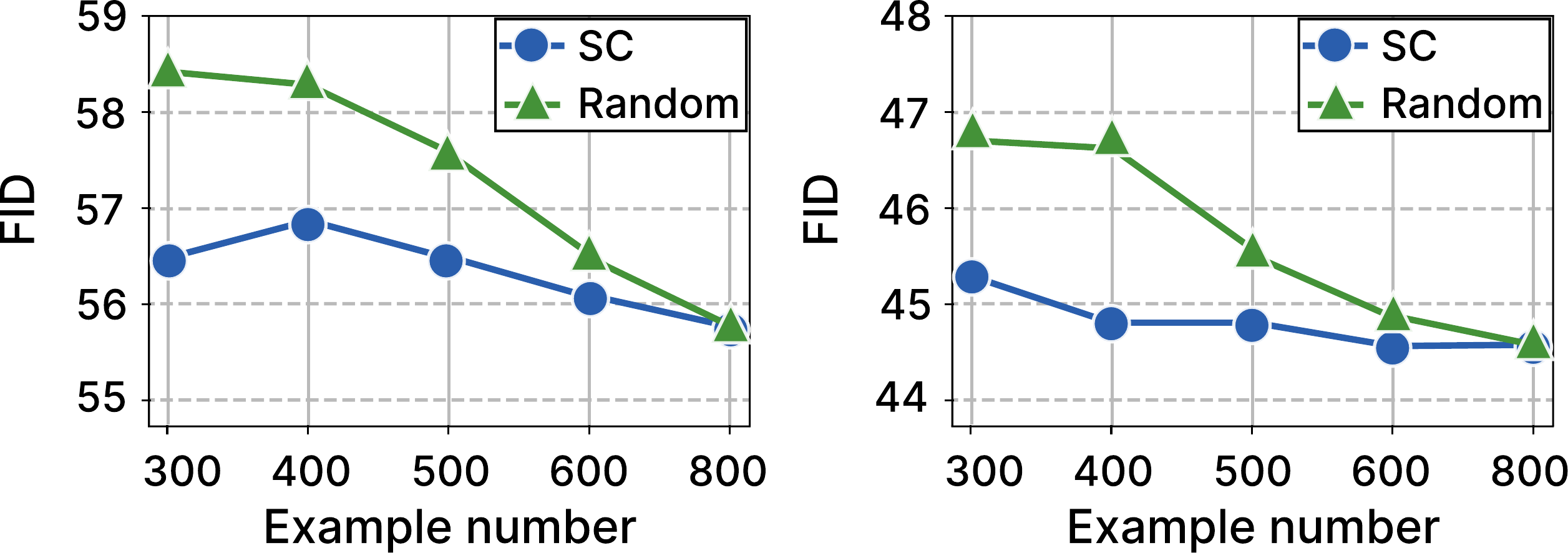}  
 \caption{ \textbf{Comparisons of Data Selection Rule.}  Prompts for the \emph{left} and \emph{right} figures are \texttt{old person} and \texttt{put on a pair of sunglasses}, respectively.} \label{fig:sc_selection} 
\end{figure}
\textbf{Analysis of Cluster Number of Data Selection.}
To investigate our data sampling rule SC for obtaining training samples (proposed in Sec.~\ref{sec:sc} to reduce the number of training data), we compare it with the random sampling method. Random sampling is implemented as shuffling the training data randomly and only accessing the first $K$ examples as training data.  The comparisons are conducted with different numbers of training samples $K$. We show the results in Fig.~\ref{fig:sc_selection} and can draw the following observations. First, SC provides better FID performance than random sampling in all scenarios, indicating the effectiveness of our sampling method by enriching data diversity. Second, the cluster number, \ie the number of target training samples, influences the SC performance to some extent. More training examples (clusters) do not necessarily lead to better performance. For instance, on the \texttt{old person} concept, a cluster number of 300 provides a better FID performance than setting the cluster number as 400. Furthermore, SC can work for a wide range of different number of training samples by providing models with good FID performance. 
\section{Conclusion}
This paper addresses the growing demand for efficient on-device image editing by introducing a novel research direction, that is the efficient training of efficient GAN models via distilling the large-scale text-to-image diffusion models with data distillation. 
The proposed framework, E$^2$GAN, incorporates a hybrid training pipeline that can efficiently adapt a pre-trained text-conditioned GAN model, which has real-time inference speed on mobile devices, to different concepts, while significantly mitigating computational and storage demands. The framework includes the construction of a base GAN model trained from various diffusion models, enabling fine-tuning for new concepts, an effective trainable parameter reduction approach, and a similarity clustering-based training data reduction method. 
Extensive experimental results validate the effectiveness of E$^2$GAN. We hope our work can shed light on how to democratize the diffusion models into efficient on-device computing.


\section*{Impact Statement}
Real-time on-device image generation with current large-scale diffusion models is still challenging. This work proposes an innovative approach to this purpose, especially in the image domain.
We leverage the data distillation approach to train lightweight GAN models on paired data prepared by large-scale text-to-image diffusion models.
In addition, we introduce an innovative architecture with attention blocks that are more efficient and can be easily adapted to new concepts with higher performance.
By saving the required tunable parameters and selecting only a small portion of data during fine-tuning, we accelerate the transfer learning process without sacrificing image quality.
Our work provides an effective way to leverage both the high-generating quality of large foundation models and the fast-generating speed of lightweight networks to enable real-time on-device image generation with high fidelity.

\textbf{Limitations.}
Generating high-quality images using diffusion models can be challenging for diverse prompts, which in turn restricts the expansion of our training datasets. Moreover, utilizing diffusion models for data collection remains an expensive endeavor. Developing efficient techniques to rapidly construct well-paired and high-quality datasets from diffusion models would greatly enhance the training of E$^2$GAN.

\textbf{Broader Impacts.}
Real-time high-quality image generation can find many fantastic applications including popular entertainment and artistic creation.
However, the widespread availability and power of these tools also pose significant challenges. Misuse and abuse of image generation models can lead to issues such as the creation of deepfakes, misleading media, and other forms of digital deception. 
Restricting abuse and misuse of powerful models with more supervision by the public or legal control will enhance the beneficial outcomes of these models and maximize the interest we could gain from them.
\nocite{langley00}

\bibliography{example_paper}
\bibliographystyle{icml2024}

\newpage
\appendix
\onecolumn
\section{Overall Algorithm for LoRA Rank Search } \label{appendix:algorithim}
\begin{algorithm}[htbp]
\SetAlgoLined
\SetNoFillComment
\textbf{Input}: Model with $I$ crucial layers, $K$ sampled concepts, training epochs $e$, upper threshold $\{\tau_i\}_{i=1}^{I}$.\\
\textbf{Output}: The rank $\{r_i^*\}_{i=1}^I$. \\
\textbf{Initialize:} $\{r_i^*\}_{i=1}^I \leftarrow \{1\}_{i=1}^I$ \\
\For{$k = 1,\ldots,K$}{
Get the concept $\mathbf{c}_k$ and paired dataset $\{(\tilde{\mathbf{x}}^{c_k}, \mathbf{x})\}$ for the concept \\
$fid \leftarrow \infty$ \\
$new\_fid \leftarrow \infty$ \\
\While{$\exists r_{i} < \tau_i$ and $new\_fid \leq fid$}{
    $\{r_i\}_{i=1}^I \leftarrow \{\min(2*r_i, \tau_i)\}_{i=1}^I$ \\
    Train $\{\theta_i^A, \theta_i^B\}_{i=1}^I$ with the rank $\{r_i\}_{i=1}^I$ for $e$ epochs on the training set of $\{(\tilde{\mathbf{x}}^{c_k}, \mathbf{x})\}$ \\
    $fid\leftarrow new\_fid$ \\
    Evaluate the FID score $new\_fid$ with current model weights on the test set of $\{(\tilde{\mathbf{x}}^{c_k}, \mathbf{x})\}$ \\
}
\If{$\{r_i\}_{i=1}^I > \{r_i^*\}_{i=1}^I$}{$\{r_i^*\}_{i=1}^I \leftarrow \{r_i\}_{i=1}^I$}
}
 \caption{LoRA rank search in Sec. \ref{sec:rank_for_lora}}
 \label{alg:lora_search}
\end{algorithm} 
We show the overall algorithm for LoRA Rank Search in Algorithm \ref{alg:lora_search}. For each concept in the $K$ sampled concepts, we start by assigning $r_i$ as 1 for each crucial layer $i$, and upscale the rank for every $e$ epochs by doubling the rank value, until $r_i$ reaches the upper threshold $\tau_i$ for the layer $i$. We evaluate the FID performance at the end of each $e$ training epochs. If the performance saturates, the rank value from the best FID performance setting is returned as the rank for the concept. Typically, a larger rank can provide more model capability. Thus, the largest returned rank among the $K$ selected concepts is viewed as $r^*$ for the future use of a new concept.

\section{More Implementation Details}
\subsection{Details for Diffusion Model}
We apply most recent diffusion-based image editing models to create paired datasets, which include Stable Diffuison (SD)~\citep{rombach2022high}, Instruct-Pix2Pix (IP2P)~\citep{brooks2022instructpix2pix},  Null-text inversion (NI)~\citep{mokady2022null}, ControlNet~\citep{zhang2023adding}, and Instruct Diffusion \citep{geng2023instructdiffusion}. 
For all these models, we use the checkpoints or pre-trained weights reported from their official websites\footnote{SD v1.5:~\url{https://huggingface.co/runwayml/stable-diffusion-v1-5}, IP2P:~\url{http://instruct-pix2pix.eecs.berkeley.edu/instruct-pix2pix-00-22000.ckp}, NI:~\url{https://huggingface.co/CompVis/stable-diffusion-v1-4}, ControlNet:~\url{https://huggingface.co/lllyasviel/ControlNet/blob/main/models/control_sd15_normal.pth}, InstructDiffusion:~\url{https://github.com/cientgu/InstructDiffusion}.}.

More specifically, for SD, 
the strength, guidance scale, and denoising steps are set to $0.68$, $7.5$, and $50$, respectively. 
For IP2P, 
images are generated with $100$ denoising steps using a text guidance of $7.5$ and an image guidance of $1.5$.
For NI, 
each image is generated with $50$ denoising steps and the guidance scale is $7.5$. The fraction of steps to replace the self-attention maps is set in the range from $0.5$ to $0.8$ while the fraction to replace the cross-attention maps is $0.8$. The amplification value for words is $2$ or $5$, depending on the quality of the generation. For ControlNet, 
the control strength, normal background threshold, denoising steps, and guidance scale are $1$, $0.4$, $20$, and $9$, respectively. For Instruct Diffusion, the denoising steps, text guidance, and image guidance are set as $100$, $5.0$, and $1.25$, respectively.  


\subsection{Hyperparameters in LoRA Rank Search}
During the process of searching LoRA rank, the rank $r_i$ for each crucial layer $i$ is upscaled once for every $e$ epochs until $r_i$ reaches the upper threshold $\tau_i$ for the layer $i$. In the experiments, $e$ is set as $10$. The rank threshold $\tau_i$ is determined by the size of the layer. More specifically, the crucial layers include: (1) four CONV-based upsampling layers with the shape as ${[3,64,7,7],[64,128,3,3],[128,256,3,3]}$, and ${[256,256,3,3]}$; (2) four corresponding downsampling layers by transpose CONV with the same set of weight shape as upsampling; and (3) transformer blocks with projection matrices $q$,$k$,$v$ with shape as ${[256,256]}$, and multi-layer perceptron (MLP) module with shape as ${[2048,256]}$ and ${[256,1024]}$. Based on the weight size, the rank threshold $\tau$ is set as $1$, $4$, $16$, and $32$ for the four upsampling/downsampling layers, respectively, and $1$ for the layers in the transformer block. After the search process, the suitable rank is determined as $1$, $4$, $8$, $8$ for the four upsampling/downsampling layers.

\subsection{Details for the Concept Setting}
    The 20 random concepts in Tab. \ref{tab:base_model_influence} include \texttt{Leonardo da Vinci painting}, \texttt{Gouache, Abstract Murals}, \texttt{Pointillist Portraits}, \texttt{Young person}, \texttt{Op Art}, \texttt{Sand Art}, \texttt{Cubist Makeup}, \texttt{Romanticism}, \texttt{Futurist Portraits}, \texttt{Hulk}, \texttt{Documentary Photography}, \texttt{Cubist Portraits}, \texttt{Pale person}, \texttt{Typography Art}, \texttt{Picasso painting}, \texttt{Photorealistic Portraits}, \texttt{Black and White Photography}, \texttt{Quilting}, \texttt{Batman}. The 30 evaluation concepts in Tab. \ref{tab:comparison} include: \texttt{Albino person}, \texttt{Angry person}, \texttt{Blond person}, \texttt{Old person}, \texttt{Grey hair}, \texttt{Put on sunglasses}, \texttt{Tan person}, \texttt{Burning man}, \texttt{Abstract Expressionist Makeup}, \texttt{Watercolor painting}, \texttt{Screen printing}, \texttt{Silver Sculpture}, \texttt{Vincent van Gogh style}, \texttt{Paul Gauguin painting}, \texttt{Henri Matisse paintings}, \texttt{Jacob Lawrence painting}, \texttt{Chinese Ink painting, Oldtime photo}, \texttt{Low Quality photo}, \texttt{Green Lantern}, \texttt{White Walker}, \texttt{Hercule Poirot}, \texttt{Ghost Rider}, \texttt{Catwoman}, \texttt{Harley Quinn}, \texttt{Chewbacca from Star Wars}, \texttt{Obi-wan Kenobi}, \texttt{Zombie}, \texttt{Gamora}, \texttt{Draco Malfoy}. The concepts selected by our approach in base generator construction as described in Sec. 4.2.1 include \texttt{Abstract Art}, \texttt{Bleeding Person}, \texttt{Burning Person}, \texttt{Comic}, \texttt{Leonardo da Vinci painting}, \texttt{Frida Kahlo painting}, \texttt{Hulk}, \texttt{Joker}, \texttt{Low Quality photo}, \texttt{Manga}, \texttt{Miro painting}, \texttt{Amedeo Modigliani painting}, \texttt{Monet painting}, \texttt{Ancient Egypt Monumental}, \texttt{Mummy}, \texttt{Munch art}, \texttt{Picasso painting}, \texttt{Pink hair}, \texttt{Pop art}, \texttt{Sketch}, \texttt{Sleeping person}, \texttt{Ukiyo-e style}, \texttt{Wax figure}, \texttt{Young person}. For the FFHQ dataset, there are 260 concepts in total, where 30 concepts are used for diverse fine-tuning purposes. The selected 20 concepts are obtained by K-means clustering with the remaining 230 concepts. For the Flickr Scenery dataset, there are 20 concepts in total, where 10 concepts are used for pretraining and the other 10 concepts are for the diverse fine-tuning purpose. 

\section{More Analysis for the Efficient Image-to-Image Model}
\subsection{Effectiveness of Model Architecture}
\begin{table}[htbp]
\centering
\caption{FID comparison between E$^2$GAN model architecture (3RB+1TB) and pix2pix (9RB) under the setting of training-from-scratch.} \label{tab:model_arch_effectiveness}
\scalebox{0.9}{
\begin{tabular}{c|c|c}
\Xhline{0.2ex}
\textbf{Concept}      & \cellcolor[gray]{.9} \textbf{E$^2$GAN (3RB+1TB)} & \textbf{Pix2pix (9RB)} \\ \hline
\texttt{Angry person} & \cellcolor[gray]{.9}\textbf{49.56}              & 55.16         \\
\texttt{Pale person}  & \cellcolor[gray]{.9}\textbf{42.65}              & 49.14         \\
\texttt{Tan person}   & \cellcolor[gray]{.9}\textbf{42.47}             & 51.37         \\
\texttt{Young person} & \cellcolor[gray]{.9}\textbf{51.27}              & 56.10         \\ \Xhline{0.2ex}
\end{tabular} }

\end{table}

Here we further show the effectiveness of our efficient model architecture design in complementary to the results in Sec.~\ref{sec:base_model_con}.
We compare our 3RB+1TB design against the 9RB design used in pix2pix for several concept settings. The results are shown in Tab.~\ref{tab:model_arch_effectiveness} with both models trained on the entire training set of $800$ samples.
From this, we can see that the 3RB+1TB design can reach higher FID with fewer parameters and FLOPs (as in Tab.~\ref{tab:size_flops_latency}).
For instance, a 3RB+1TB model in the target concept domain of \texttt{pale person} has a FID as $42.65$, decreasing the FID value by $6.49$ compared to the 9RB model of pix2pix.

\begin{table}[h]
    \centering
    \caption{Quantitative results (FID) on the AFHQ dataset for different concepts.}
    \begin{tabular}{lccc}
        \Xhline{0.2ex}
        \textbf{Model} & \textbf{Pix2pix} & \textbf{Co-Mod-GAN} & \cellcolor[gray]{.9} \textbf{Ours} \\
        \hline
        \texttt{Cat to fox} & 39.78 & 42.67 & \cellcolor[gray]{.9} \textbf{36.60} \\
        \texttt{Cat to ocelot} & 30.72 & 33.64 & \cellcolor[gray]{.9} \textbf{29.51} \\
        \texttt{Vincent van Gogh style} & 67.11 & 66.83 & \cellcolor[gray]{.9} \textbf{64.09} \\
        \texttt{Charcoal drawing} & 28.01 & 28.58 & \cellcolor[gray]{.9} \textbf{25.90} \\
        \texttt{Pop art} & 112.58 & 132.78 & \cellcolor[gray]{.9} \textbf{110.28} \\
        \Xhline{0.2ex}
    \end{tabular}
    \label{tab:afhq_results}
\end{table}
\begin{table}[]
    \centering
    \caption{Quantitative results on conventional benchmarks for paired data.}
    \begin{tabular}{lccc}
        \Xhline{0.2ex}
        \textbf{Model} & \textbf{Facades (FID)} & \textbf{Cityscapes (mIoU)} & \textbf{Edges $\rightarrow$ Shoes (FID)} \\
        \hline
        Pix2pix & 126.65 & 42.06 & 24.18 \\
        Co-Mod-GAN & 136.72 & 35.62 & 38.50 \\
        GAN Compression & - & 41.71 & 25.76 \\
        \rowcolor[gray]{.9} \textbf{Ours} &  \textbf{121.89} & \textbf{43.20} &  \textbf{24.03} \\
\Xhline{0.2ex}
    \end{tabular}
    \label{tab:conventional_paired_results}
\end{table}

\begin{table}[]
    \centering
    \caption{Quantitative results on the unpaired dataset.}
    \begin{tabular}{lc}
        \Xhline{0.2ex}
        \textbf{Model} & \textbf{Horse2Zebra (FID)} \\
        \hline
        CycleGAN & 74.04 \\
        CUT & 45.76 \\
        GAN Compression & 64.95 \\
        GAN Slimming & 86.09 \\
        \rowcolor[gray]{.9} \textbf{Ours} & \textbf{44.12} \\
        \Xhline{0.2ex}
    \end{tabular}
    \label{tab:unpaired_results}
\end{table}

To demonstrate the generalization ability of our model architecture, we further include the results on the AFHQ dataset. We follow the same pipeline as in the main results. We use $1,000$ images in the AFHQ dataset to generate paired data with diffusion models. The base generator is trained on three concepts including cat to serval, watercolor painting, and chalk art. The performance is evaluated on five concepts. We provide quantitative results in Tab. \ref{tab:afhq_results} and the generated images in Fig. \ref{fig:afhq}. The results show that our method performs better than baseline methods, indicating the generalization ability.

We further conduct experiments on other image-to-image translation tasks other than diffusion model distillation to further show the effectiveness of our model architecture design. We use conventional paired benchmark datasets, including sketch->shoes, facades, cityscapes, and conventional unpaired benchmark dataset, such as horse2zebra. We provide quantitative results in Tab. \ref{tab:conventional_paired_results} and \ref{tab:unpaired_results}.
From the results, we can observe that our design achieves better performance with higher FID and lower mIoU.
\begin{figure*}[]
\centering{
\includegraphics[width=0.7 \linewidth]{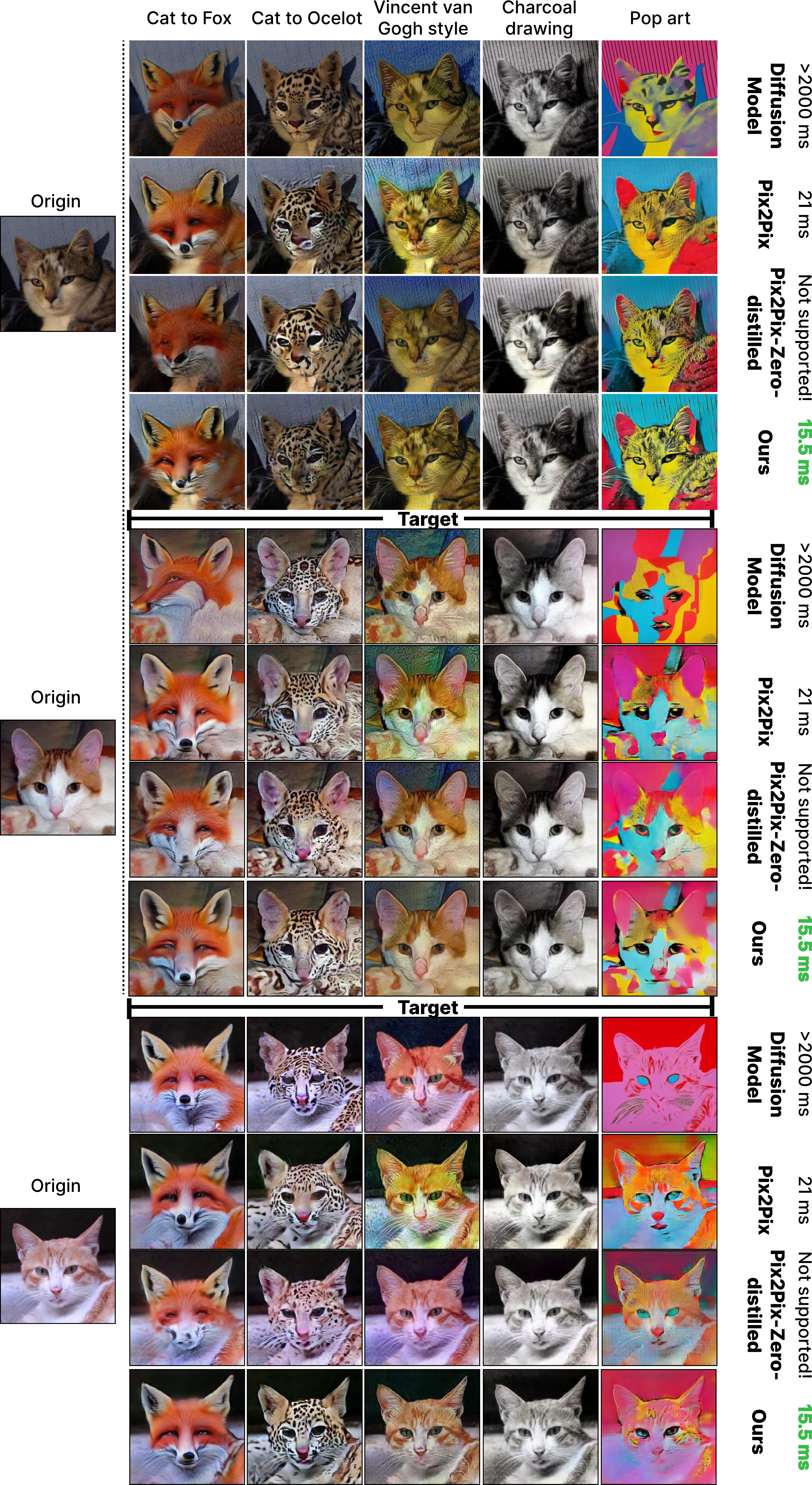}
}
\caption{
\textbf{Qualitative comparisons} on various tasks. The \emph{leftmost} column shows original images and the remaining columns present the corresponding synthesized images in the target concept domain.
}
\label{fig:afhq} 
\end{figure*}

\subsection{Sampling Operations for Transformer Block}
As mentioned in Sec.~\ref{sec:base_model_con}, we apply a downsampling operation with a CONV layer to halve the feature map size before sending it into the transformer block, and use an upsampling layer implemented by transpose CONV operation to recover the feature map size for the following operations to reduce the amount of computations. We conduct another set of experiments on the Flicker Scenery dataset to see if these sampling operations can be replaced by pooling and unpooling operations, such that a smaller model size can be reached. We first train these two models on the selected prompts to get the base model. Then, we fine-tune the entire model with all the training data for a new concept. The comparison results are shown in Tab.~\ref{tab:max_pool}. From the results, we can observe that though applying pooling operations can reduce the number of parameters from the base model by $1.2$M, the FID performance becomes much worse. Thus, we use CONV operation instead of pooling to tackle the feature map reduction and recovery for the transformer block. 
\begin{table}[htbp]
\centering
\caption{FID performance of replacing the downsampling and upsampling layers for the transformer block with Max Pool and Max Unpool operations.} \label{tab:max_pool}
\scalebox{0.8}{
\begin{tabular}{c|c|c}
\Xhline{0.2ex}
         \textbf{ Operation}             & \cellcolor[gray]{.9} \begin{tabular}[c]{@{}c@{}}     
      \textbf{\textbf{CONV + }} \\ 
      \textbf{\textbf{transpose CONV}} 
      \end{tabular} &  \begin{tabular}[c]{@{}c@{}}     
      \textbf{\textbf{Max Pool + }} \\ 
      \textbf{\textbf{Max Unpool}} 
      \end{tabular} \\ \hline
       \diagbox{\textbf{Concept}}{\textbf{Model Size}}              & \cellcolor[gray]{.9}7.1M                  & 5.9M                  \\ \hline
\texttt{Forest in the dark}     & \cellcolor[gray]{.9} \textbf{121.60}               & 190.05                \\
\texttt{Impressionism painting} & \cellcolor[gray]{.9} \textbf{88.52}                & 135.96                \\
\texttt{Forest in the autumn}   & \cellcolor[gray]{.9} \textbf{88.82}                 & 141.29                \\ \Xhline{0.2ex}
\end{tabular}}

\end{table}

\section{More Ablation Analysis for the Base Model}
\subsection{Pre-train with Multiple Concepts for Conventional GAN Training}
We investigate if conventional GAN training such as pix2pix can benefit from fine-tuning a pre-trained base model, as leveraged in E$^2$GAN. To verify this, we follow the same step as E$^2$GAN to pre-train pix2pix with the selected $7$ prompts/datasets on the Flicker Scenery dataset. Then, the base model is fine-tuned to adapt to other concepts. The results in Tab.~\ref{tab:pretrain_pix2pix} show that pix2pix does not gain much benefits from pre-training. Moreover, the performance becomes even worse, such as for the concept \texttt{Vangogh style} (FID degrades from $138.77$ to $151.20$ with a pre-trained base model). The results indicate that with our efficient architecture design, our base model possesses the
capability of more general features and 
representations when trained on multiple concepts. The transformer block with self-attention modifies the image with a better holistic understanding and
the cross-attention module takes the information from the given target concept. Thus, our method allows the new concept to better leverage existing knowledge, which is not possessed by prior methods.
\begin{table}[htbp]
\centering
\caption{FID performance of fine-tuning from a pre-trained base model for pix2pix on Flicker Scenery dataset. } \label{tab:pretrain_pix2pix}
\scalebox{0.9}{
\begin{tabular}{c|cc|c}
\Xhline{0.2ex}
\textbf{Method}             & \multicolumn{2}{c|}{\textbf{Pixpix}}                                        & \cellcolor[gray]{.9} \textbf{E$^2$GAN}              \\ \hline
        \diagbox{\textbf{Concept}}{\textbf{Pretrain}}            & \multicolumn{1}{c|}{\cmark} & \xmark & \cellcolor[gray]{.9} \cmark \\ \hline
\texttt{Vangogh style}     & \multicolumn{1}{c|}{151.2}                 & 138.77                & \cellcolor[gray]{.9} \textbf{117.41}                \\
\texttt{Add blossoms}        & \multicolumn{1}{c|}{157.76}                & 150.96                & \cellcolor[gray]{.9}\textbf{146.42}                \\ 
\texttt{Forest in the winter}        & \multicolumn{1}{c|}{119.31}                & 122.35               & \cellcolor[gray]{.9}\textbf{119.15}      \\          
\Xhline{0.2ex}
\end{tabular}}

\end{table}

\subsection{Autoencoder as Pre-trained Base Model}
\begin{table}[htbp]
\centering
\caption{The FID performance of using autoencoder as the pre-trained base model.} \label{tab:auto_encoeder}
\scalebox{1}{
\begin{tabular}{c|c|c}
\Xhline{0.2ex}
  \diagbox{\textbf{Base model}}{\textbf{Concept}}    & \begin{tabular}[c]{@{}c@{}}     
      {\texttt{Angry}} \\ 
      {\texttt{person}} 
      \end{tabular} & \begin{tabular}[c]{@{}c@{}}     
      {\texttt{White}} \\ 
      {\texttt{walker}} 
      \end{tabular} \\ \hline
Auto-encoder & 110.35 & 80.43 \\
\texttt{Old person} & 54.48 & 51.99 \\
\rowcolor[gray]{.9} \textbf{Ours} & \textbf{54.27} & \textbf{40.18} \\ \Xhline{0.2ex}
\end{tabular}}
\end{table}
In E$^2$GAN, we first train the GAN model with multiple diverse concepts to get a pre-trained base model, and then fine-tune it to other concepts. 
We have shown multiple base model settings in Sec.~\ref{sec:ablation_analysis}.
One may wonder if the pre-trained base model can be chosen as an auto-encoder, \eg the base model encodes the input data into a lower-dimensional representation and then decodes it back into the original data, instead of being trained on other concepts.
To verify this, we conduct experiments by first training an auto-encoder on the original images in the subset of FFHQ~\citep{karras2019style} with only the $\ell_1$ loss in Eq.~\ref{eq:loss}, then fine-tune the auto-encoder following the same method as fine-tuning a pre-trained GAN.
The results are compared in Tab.~\ref{tab:auto_encoeder}.
We find that auto-encoder is not comparable as fine-tuning a GAN trained on a single concept as \texttt{old person}, not to mention our base model that is pre-trained on multiple concepts.
For instance, for the target style \texttt{angry person}, tuning from a base model pre-trained to generate \texttt{old person} can give an FID as low as $54.48$, yet tuning from the auto-encoder results in a much worse FID of $110.35$. This might due to the simplicity of the auto-encoder, which only needs to generate the original image and does not necessarily include other semantic information, either coarse-grained global features, or fine-grained local details.
In contrast, the GAN models include more information like texture or color, during training.
From this observation, in E$^2$GAN, we adopt a model pre-trained on several concepts instead of using auto-encoder as the base model.

\subsection{Removing Cross-Attention During Fine-Tuning}
\begin{table}[]
    \centering
    \caption{Quantitative results for different concepts.}
    \begin{tabular}{lcc}
        \Xhline{0.2ex}
        \textbf{Concept} & \textbf{Remove Cross-Attention} &  \cellcolor[gray]{.9} \textbf{Ours} \\
        \hline
        \texttt{Vincent van Gogh} & 90.31 &  \cellcolor[gray]{.9} \textbf{71.82} \\
        \texttt{Blond Person} & 59.78 &  \cellcolor[gray]{.9} \textbf{48.01} \\
        \texttt{White Walker} & 55.43 &  \cellcolor[gray]{.9} \textbf{40.18} \\
        \Xhline{0.2ex}
    \end{tabular}
    \label{tab:cross_attention_results}
\end{table}
We also considered removing the cross-attention layers during the fine-tuning to save the computation, yet the image generation ability is degraded obviously. We provide the FID evaluations of removing the cross-attention on the FFHQ dataset across several different concepts in Tab. \ref{tab:cross_attention_results}. The rationale behind the results is that the cross-attention takes both the text information and image feature information as input to compute the output feature map for the next building block. Directly removing the cross-attention block from the base model during the fine-tuning phase will make the feature map have different meanings, thus influencing the image generation quality.


\section{Ablation on the Influence of Longer Training Time}
\begin{table}[htbp]
\centering
\caption{The FID comparison between training E$^2$GAN for 100 epochs and 200 epochs.} \label{tab:longer_train}
\scalebox{0.95}{
\begin{tabular}{c|c|c}
\Xhline{0.2ex}
\textbf{Concept}              & \cellcolor[gray]{.9} \begin{tabular}[c]{@{}c@{}}     
      \textbf{\textbf{Train 100}} \\ 
      \textbf{\textbf{epochs}} 
      \end{tabular} & \begin{tabular}[c]{@{}c@{}}     
      \textbf{\textbf{Train 200}} \\ 
      \textbf{\textbf{epochs}} 
      \end{tabular} \\ \hline
\texttt{Forest in the dark}   & \cellcolor[gray]{.9} 115.32          & 114.17          \\
\texttt{Oil painting }        & \cellcolor[gray]{.9} \textbf{110.87}          & 111.93          \\
\texttt{Forest in the spring} & \cellcolor[gray]{.9} \textbf{122.77}          & 124.91          \\ \Xhline{0.2ex}
\end{tabular}}
\end{table}

E$^2$GAN greatly saves training time compared to conventional GAN training while maintaining good image synthesis ability.  To see if training longer can lead to better performance, we add further experiments to increase the training time by doubling the training epochs. The results can be found in Tab. \ref{tab:longer_train}. The reported FID is evaluated on the model weights obtained at the end of training.  The results show that training longer will not bring obvious performance improvements for E$^2$GAN, but leads to more computation cost. The results indicate that our efficient E$^2$GAN is able to reach good performance with fewer epochs compared to conventional GAN training. 

\section{Diffusion Model Data Challenge}
\begin{figure*} [htbp]
     \centering
     \includegraphics[width=1\textwidth]
      {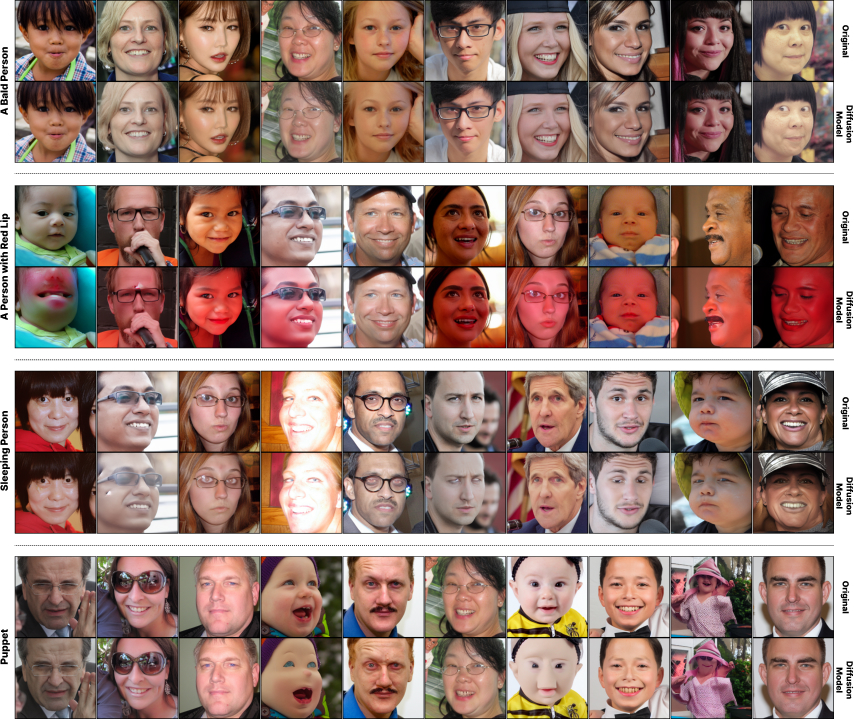}  
     \caption{{Examples of the cases that diffusion models do not work. For each group of images, the target concept is shown on the left, the first row demonstrates the original image, and the second row shows the corresponding synthesized images in the target concept domain.
     }}
    \label{fig:diffusion_prob_1}
\end{figure*}

\begin{figure*} [t]
     \centering
     \includegraphics[width=1\textwidth]{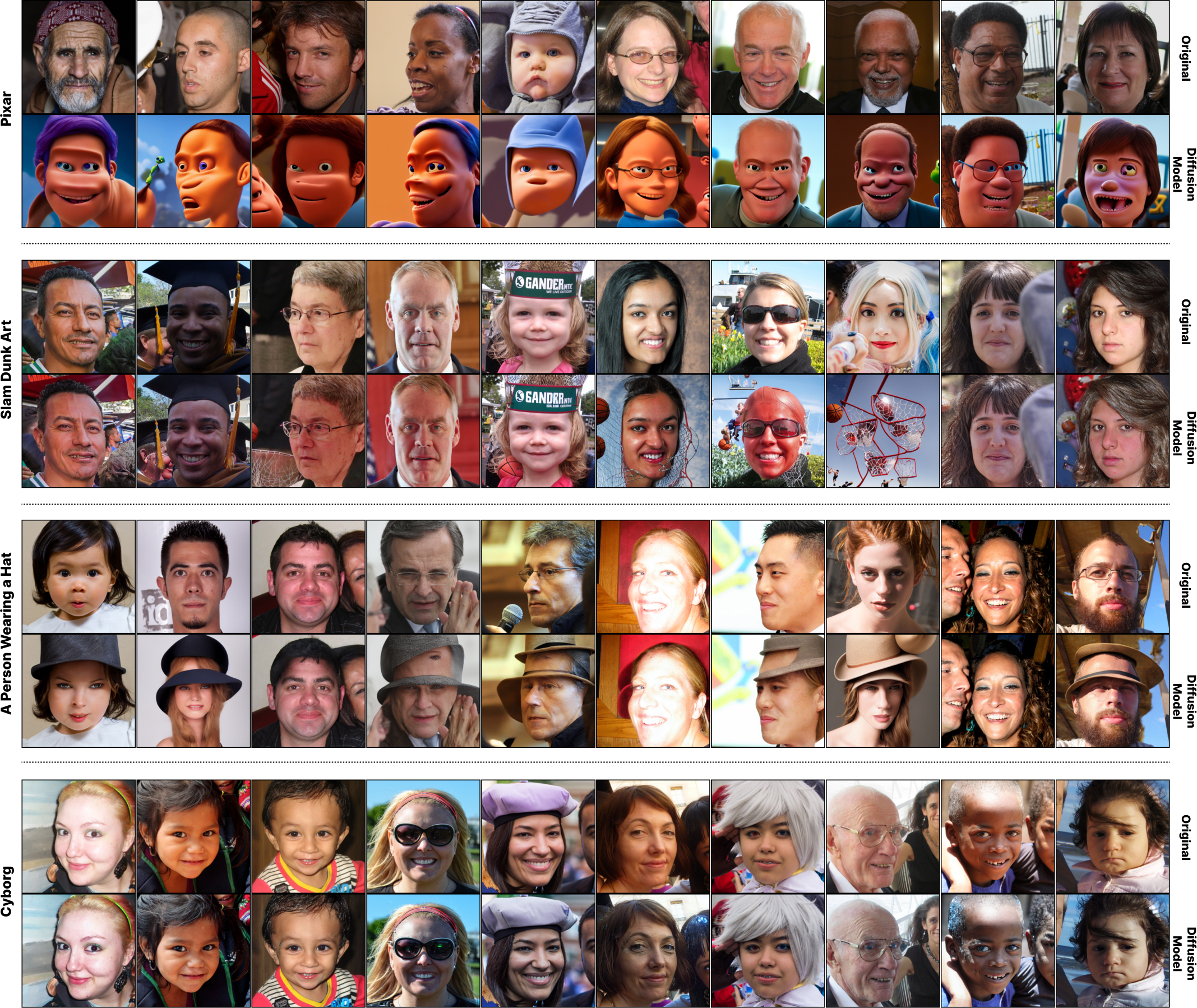}  
     \caption{Examples of the cases that diffusion models do not work. For each group of images, the target concept is shown on the left, the first row demonstrates the original image, and the second row shows the corresponding synthesized images in the target concept domain.}
    \label{fig:diffusion_prob_2}
\end{figure*}
Generating data through the diffusion models to transfer the knowledge to lightweight GAN models poses certain challenges. While text-to-image diffusion models exhibit excellent capabilities in generating high-quality images, they do not consistently perform well in all scenarios. We illustrate this by presenting some examples below as in Fig. \ref{fig:diffusion_prob_1} and Fig. \ref{fig:diffusion_prob_2}. For instance, for the concept \texttt{ A person with red lip} in Fig.\ref{fig:diffusion_prob_1}, the diffusion model (IP2P) usually turns the entire image into the red color or modifies the person in the image to a strange shape.

\section{Additional Qualitative Results}
We provide more example images generated by our approach and other baseline methods in Fig.~\ref{fig:appendix_fig_1},~\ref{fig:appendix_fig_2},~\ref{fig:appendix_fig_3},~\ref{fig:appendix_fig_4}, and~\ref{fig:appendix_fig_5}.

\begin{figure*} [htbp]
     \centering
     \includegraphics[width=1\textwidth]{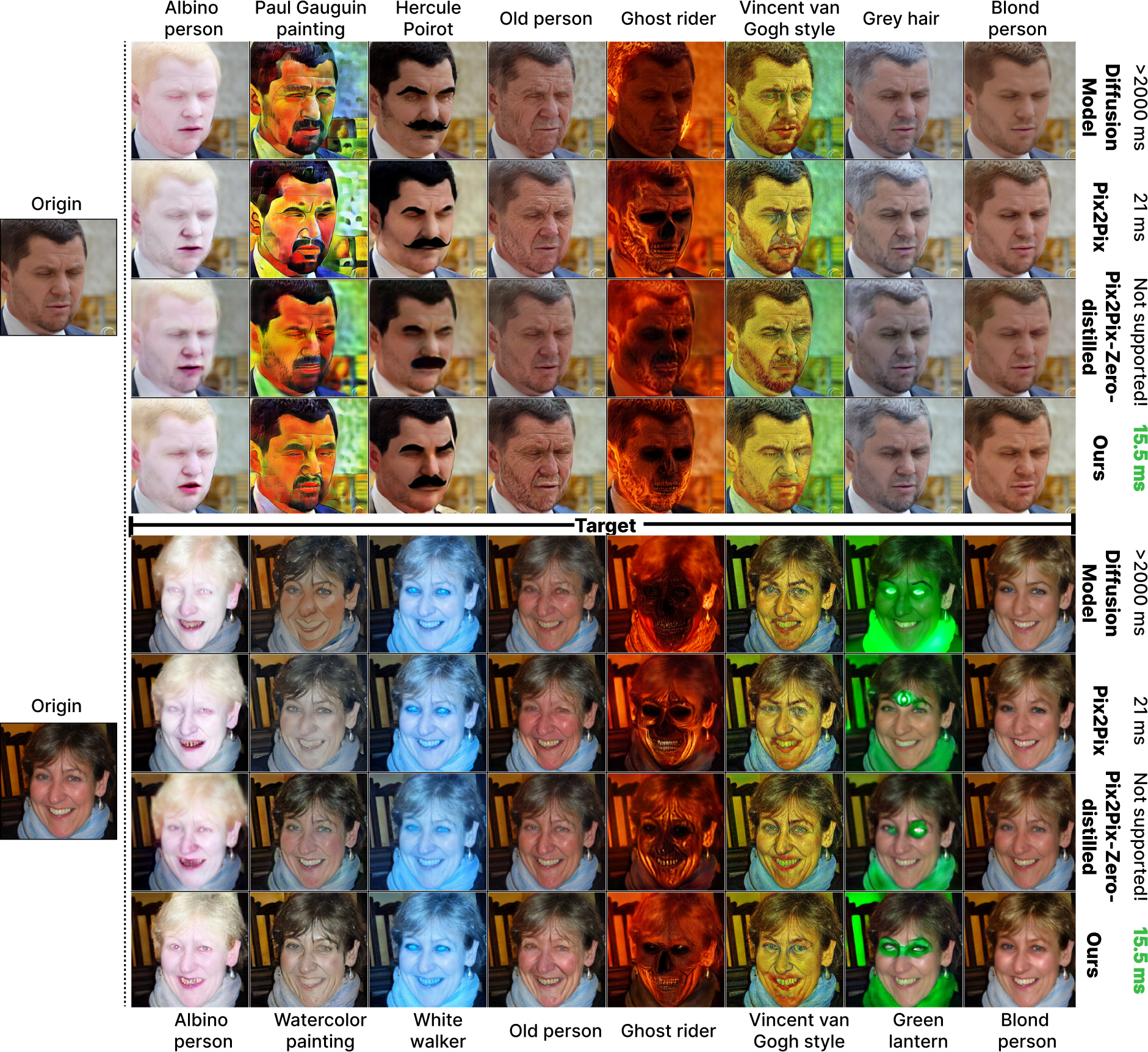}  
     \caption{{\textbf{Qualitative comparisons} on various tasks. The \emph{leftmost} column shows two original images and the remaining columns present the corresponding synthesized images in the target concept domain, where target prompts are shown at the bottom row. We provide images generated by various models.}}
    \label{fig:appendix_fig_1}
\end{figure*}

\begin{figure*} [t]
     \centering
     \includegraphics[width=1\textwidth]{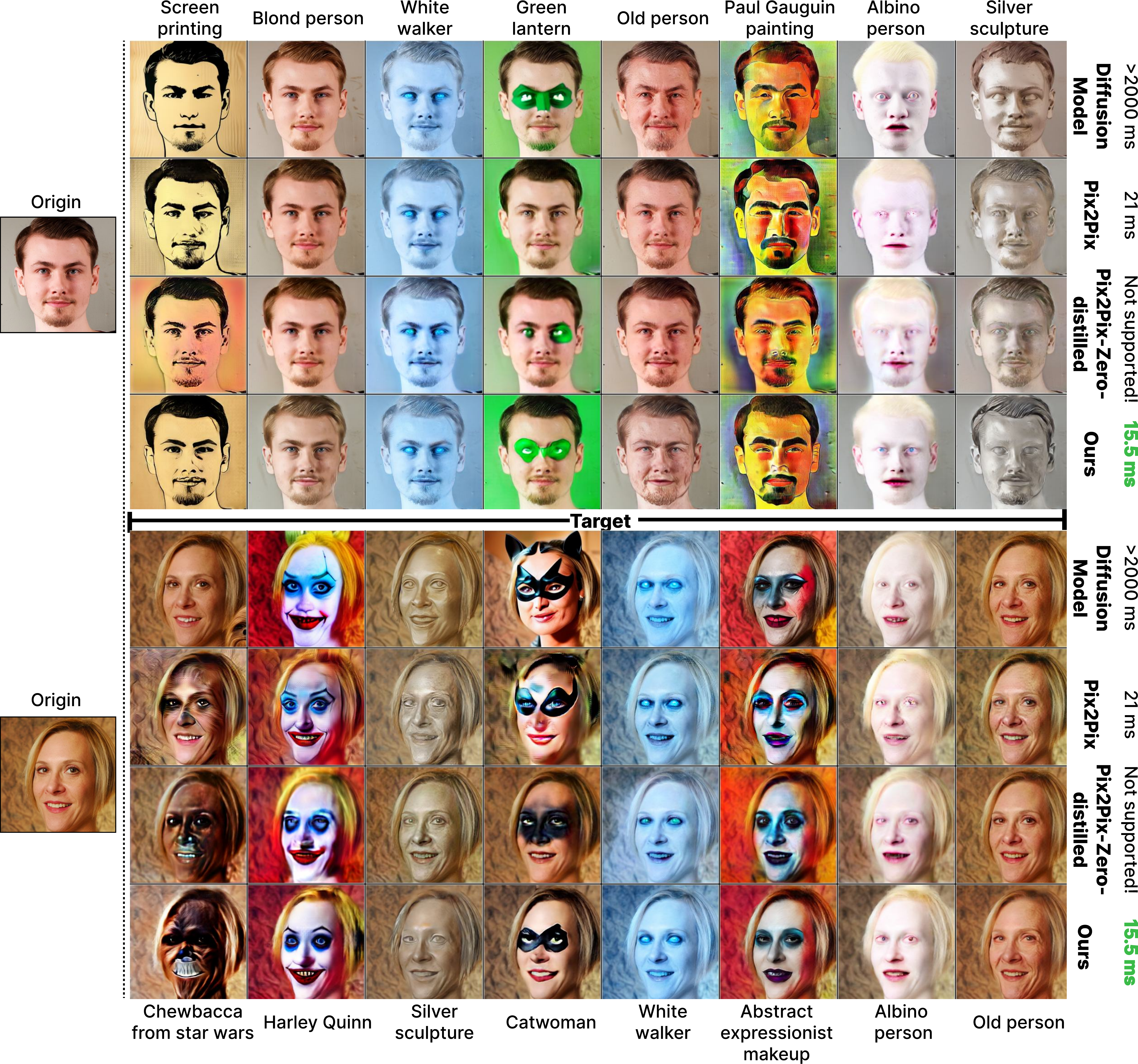}  
     \caption{{\textbf{Qualitative comparisons} on various tasks. The \emph{leftmost} column shows two original images and the remaining columns present the corresponding synthesized images in the target concept domain, where target prompts are shown at the bottom row. We provide images generated by various models.}}
    \label{fig:appendix_fig_2}
\end{figure*}

\begin{figure*} [t]
     \centering
 \includegraphics[width=1\textwidth]{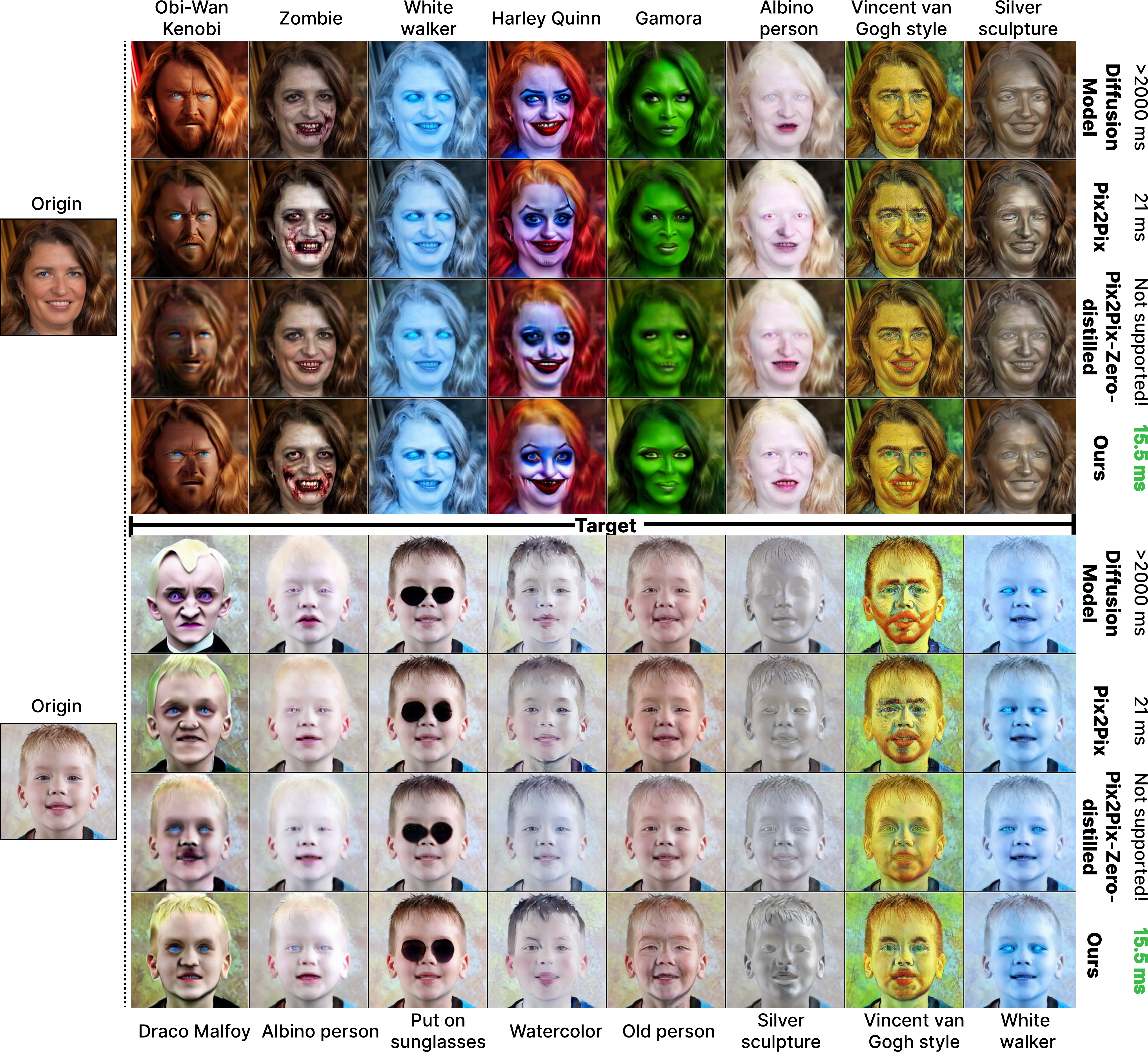}  
     \caption{{\textbf{Qualitative comparisons} on various tasks. The \emph{leftmost} column shows two original images and the remaining columns present the corresponding synthesized images in the target concept domain, where target prompts are shown at the bottom row. We provide images generated by various models.}}
    \label{fig:appendix_fig_3}
\end{figure*}

\begin{figure*} [t]
     \centering
     \includegraphics[width=1\textwidth]{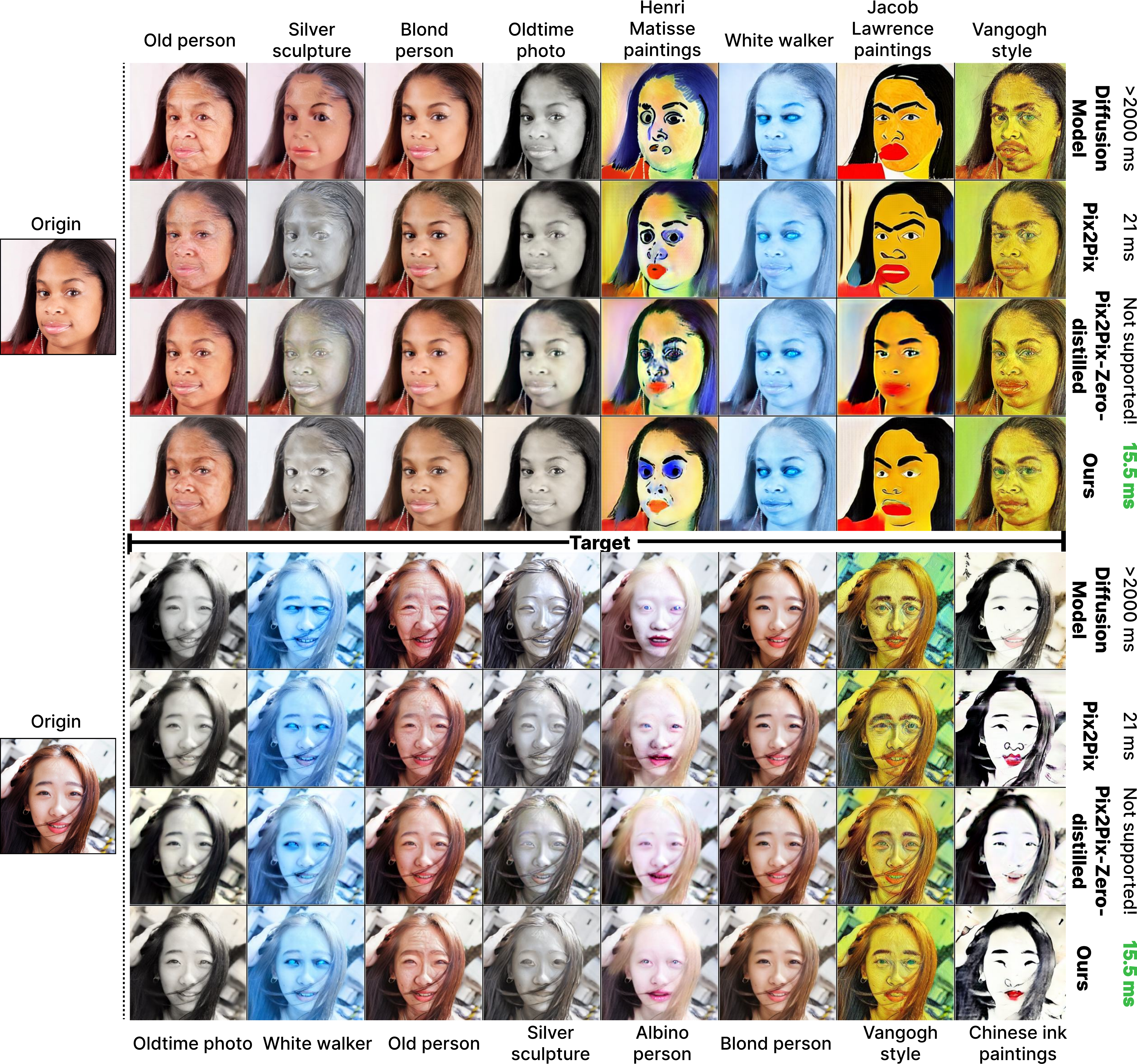}  
     \caption{{\textbf{Qualitative comparisons} on various tasks. The \emph{leftmost} column shows two original images and the remaining columns present the corresponding synthesized images in the target concept domain, where target prompts are shown at the bottom row. We provide images generated by various models.}}
    \label{fig:appendix_fig_4}
\end{figure*}

\begin{figure*} [t]
     \centering
     \includegraphics[width=1\textwidth]{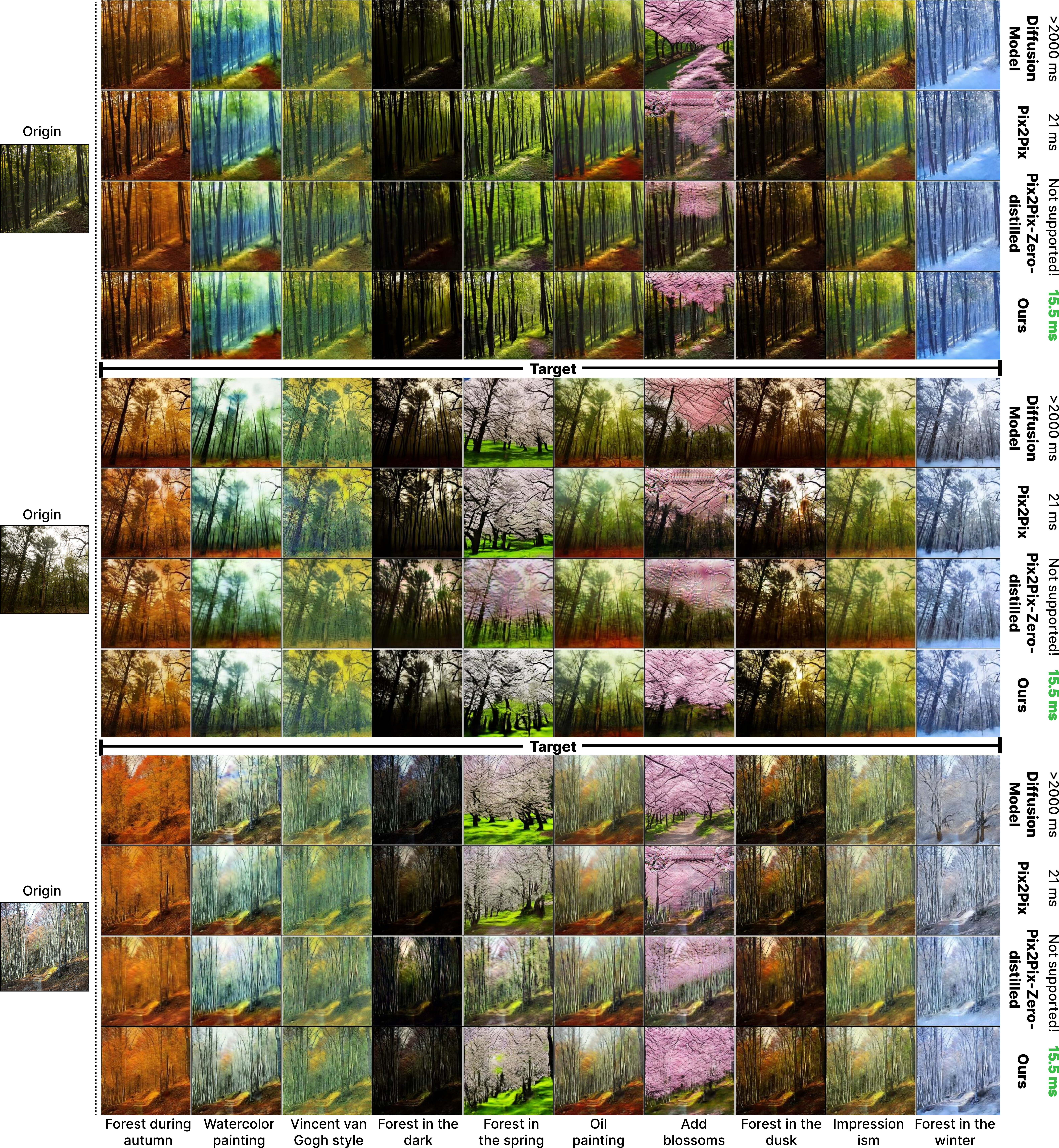}  
     \caption{{\textbf{Qualitative comparisons} on various tasks. The \emph{leftmost} column shows two original images and the remaining columns present the corresponding synthesized images in the target concept domain, where target prompts are shown at the bottom row. We provide images generated by various models.}}
    \label{fig:appendix_fig_5}
\end{figure*}


\end{document}